\documentclass[runningheads]{llncs}
\usepackage{graphicx}
\usepackage{tikz}
\usepackage{comment}
\usepackage{amsmath,amssymb} 
\usepackage{color}
\usepackage{booktabs}
\usepackage{epsfig}
\usepackage{graphicx}
\usepackage[countmax]{subfloat}
\usepackage[font=scriptsize]{subfig}
\usepackage{paralist}
\usepackage{multirow}
\usepackage{soul}
\usepackage{xcolor}
\usepackage[export]{adjustbox}
\usepackage{algpseudocode}
\usepackage[pagebackref=true,breaklinks=true,colorlinks,bookmarks=false,citecolor=blue,linkcolor=blue,urlcolor=blue]{hyperref}
\newcommand{\gb}[1]{\fcolorbox{green}{white}{#1}}
\newcommand{\rb}[1]{\fcolorbox{red}{white}{#1}}

\newcommand{\mr}[2]{\multirow{#1}{*}{#2}}

\newcommand{\fst}[1]{\textcolor{red}{\textbf{#1}}}
\newcommand{\snd}[1]{\textcolor{blue}{\underline{#1}}}

\newcommand{\ph}[1]{\paragraph{\textbf{#1}}}

\usepackage[accsupp]{axessibility}  

\begin{document}
\pagestyle{headings}
\mainmatter

\title{Learning Discriminative Shrinkage Deep Networks for Image Deconvolution
} 

\author{Pin-Hung Kuo\inst{1} \quad 
Jinshan Pan\inst{2} \quad
Shao-Yi Chien\inst{1} \quad 
Ming-Hsuan Yang\inst{3,4,5}}

\institute{$^1$Graduate Institute of Electronics Engineering, National Taiwan University \\ 
\qquad $^2$Nanjing University of Science and Technology \\ 
\qquad $^3$Google Research 
\qquad $^4$University of California, Merced 
\qquad $^5$Yonsei University 
}

\maketitle

\begin{abstract}
Most existing methods usually formulate the non-blind deconvolution problem into a maximum-a-posteriori framework and address it by manually designing a variety of regularization terms and data terms of the latent clear images. However, explicitly designing these two terms is quite challenging and usually leads to complex optimization problems which are difficult to solve. This paper proposes an effective non-blind deconvolution approach by learning discriminative shrinkage functions to model these terms implicitly. Most existing methods use deep convolutional neural networks (CNNs) or radial basis functions to learn the regularization term simply. In contrast, we formulate both the data term and regularization term and split the deconvolution model into data-related and regularization-related sub-problems according to the alternating direction method of multipliers. We explore the properties of the Maxout function and develop a deep CNN model with a Maxout layer to learn discriminative shrinkage functions to approximate the solutions of these two sub-problems directly. Moreover, the fast-Fourier-transform-based image restoration usually leads to ringing artifacts. At the same time, the conjugate-gradient-based approach is time-consuming; we develop the Conjugate Gradient Network to restore the latent clear images effectively and efficiently. Experimental results show that the proposed method performs favorably against the state-of-the-art in terms of efficiency and accuracy. Source codes, models, and more results are available at  \href{https://github.com/setsunil/DSDNet}{https://github.com/setsunil/DSDNet}.

\end{abstract}


\section{Introduction}

\label{sec:intro}

The single image deconvolution, or deblurring, aims to restore a clear and sharp image from a single blurry input image. Blind image deblurring has attracted interest from many researchers \cite{levin2011understanding,krishnan2011blind,ren2016image,pan2016blind}.
With the rapid development of deep learning, tremendous progress has been made in blind image deblurring recently \cite{kupyn2018deblurgan,tao2018scale,zhang2019deep,gao2019dynamic,suin2020spatially}. 
Since the kernel is available via blind methods, how to utilize these kernels well is still an important issue. Therefore, non-blind deconvolution has never lost the attention of researchers over the past decades \cite{richardson1972bayesian,geman1992constrained,cho2011handling,dong2020deep}.
Due to the deconvolution problem's ill-posedness, numerous methods explore the statistical properties of clear images as the image priors (e.g., hyper-Laplacian prior \cite{krishnan2009fast,levin2009understanding}) to make this problem tractable. Although using the hand-crafted image priors facilitates ringing artifacts removal, the fine details are not restored well, as these limited priors may not model the inherent properties of various latent images sufficiently. 

To overcome this problem, discriminative image priors are learned from training examples \cite{schuler2013machine,schmidt2014shrinkage,dong2018learning,ren2019simultaneous}. 
These methods usually leverage radial basis functions (RBFs) as the shrinkage functions of the image prior. However, the RBFs contain many parameters, leading to complex optimization problems. 

Deep convolutional neural network (CNN) has been developed to learn more effective regularization terms for the deconvolution problem \cite{fcn/cvpr17}. These methods are motivated by \cite{schmidt2014shrinkage} and directly estimate the solution to the regularization-related sub-problem by deep CNN models.
As analyzed by \cite{schmidt2014shrinkage}, we can obtain the solution to the regularization-related sub-problem by combining the shrinkage functions.
However, as the shrinkage functions are complex (e.g., non-Monotonic), simply using the convolution operation followed by the common activation functions, e.g., ReLU, cannot model the property of the shrinkage functions.
Given the effectiveness of the deep features, it is of great interest to learn discriminative shrinkage functions.
Therefore, if we can learn more complex shrinkage functions corresponding to the deep features, they shall surpass the hand-crafted ones in solving the regularization-related sub-problem.

We note that image restoration involves an image deconvolution step, usually taking advantage of fast Fourier transform (FFT) \cite{schmidt2014shrinkage,pan2016blind,zhang2017learning} or the Conjugate Gradient (CG) method \cite{barrett1994templates,levin2007image,dong2020deep}. 
However, the FFT-based approaches usually lead to ringing artifacts, while the CG-based ones are time-consuming.
Besides, these two methods suffer from information loss: for FFT, we lose little information when we discard the imaginary parts in the inverse real FFT; for the CG method, the iterations we execute are usually far less than the upper bound.
Therefore, developing an effective yet efficient image restoration method is also necessary.

In this paper, we develop a simple and effective model to discriminatively learn the shrinkage functions for non-blind deconvolution, which is called Discriminative Shrinkage Deep Network (DSDNet).
We formulate the data and regularization terms as the learnable ones and split the image deconvolution model into the data-related sub-problem and regularization-related sub-problem. 
As shrinkage functions can solve both these sub-problems,  and the learnable Maxout functions can efficiently approximate any complex functions, we directly learn the shrinkage functions of sub-problems via a deep CNN model with Maxout layers \cite{goodfellow2013maxout}.
To effectively and efficiently generate clear images from the output of the learned functions, we develop a fully convolutional Conjugate Gradient Network (CGNet) motivated by the mathematical concept of the CG method.
Finally, we formulate our method into an end-to-end network and solve the proposed approach by an Alternating Direction Method of Multipliers (ADMM) \cite{parikh2014proximal}.
Experimental results show that the proposed method performs favorably against the state-of-the-art ones.

The main contributions of this work are:
\begin{compactitem}
\item We propose a simple yet effective non-blind deconvolution model to directly learn discriminative shrinkage functions to model the data and regularization terms for image deconvolution implicitly.
\item  We develop an efficient and effective CGNet to restore clear images without the problems of CG and FFT.
\item The architecture of DSDNet is designed elaborately, which makes it flexible in model size and easy to be trained. Even the smallest DSDNet performs favorably against the state-of-the-art methods in speed and accuracy.
\end{compactitem}


\section{Related Work}

\label{sec:rw}
Because numerous image deconvolution methods have been proposed, we discuss those most relevant to this work. 

\ph{Statistical Image Prior-Based Methods.}
Since non-blind deconvolution is an ill-posed problem, conventional methods usually develop image priors based on the statistical properties of clear images. 
Representative methods include total variation \cite{rudin1994total,chan1998total,wang2008new,perrone2014total}, hyper-Laplacian prior \cite{levin2007image,krishnan2009fast}, and patch-based prior \cite{sun2013edge,gu2014weighted,zoran2011learning}, to name a few. 
However, these hand-crafted priors may not model the inherent properties of the latent image well; thus, these methods do not effectively restore realistic images. 

\ph{Learning-Based Methods.} 
To overcome the above limitations of the hand-crafted priors, researchers have proposed learning-based approaches, e.g., Markov random ﬁelds \cite{roth2005fields,samuel2009learning}, Gaussian mixture models \cite{zoran2011learning}, conditional random ﬁelds \cite{tappen2007learning,jancsary2012loss,schmidt2013discriminative,schmidt2015cascades}, and radial basis functions \cite{schmidt2014shrinkage,dong2018learning}.
%

The learning-based non-blind deconvolution also gets deeper with the development of neural networks.
%
%
Many methods use deep CNNs to model the regularization term and solve image restoration problems by unrolling existing optimization algorithms. For example,  Iterative Shrinkage-Thresholding algorithm \cite{zhang2018ista,xiang2021fista}, Douglas-Rachford method \cite{aljadaany2019douglas}, Half-Quadratic Splitting
algorithm \cite{li2019deep,zhang2020deep,ko2020deep,fcn/cvpr17,bigdeli2017deep,li2019blind,chen2021learning,kong2021deep}, gradient descend \cite{gong2020learning,qiu2021gradirn} and ADMM \cite{yang2016deep}.
These methods use deep CNN models to estimate the solution to the regularization-related sub-problem.
As demonstrated by \cite{schmidt2014shrinkage}, the solutions are the combinations of the shrinkage functions. 
Simply using deep CNN models does not model the shrinkage functions well since most activation functions are too simple.
Besides, most of them focus on the regularization terms yet ignore the importance of data terms.
In addition, the image restoration step in these methods usually depends on an FFT-based solution. However, using FFT may lead to results with ringing artifacts. Even though the edge taper \cite{kruse2017learning} alleviates artifacts, they are still inevitable in many scenes.

To overcome these problems, we leverage Maxout layers to learn discriminative shrinkage functions for regularization and data terms and develop the CGNet to restore images better. Furthermore, we adopt average pooling for noise level estimation and residual block for re-weights computation. In other words, we design each component according to the mathematical characteristics rather than stacking as many convolutional layers as possible, as in most previous works. 

\ph{Blind Deblurring Methods.} 
Numerous end-to-end deep networks \cite{sun2015learning,nah2017deep,tao2018scale,kupyn2018deblurgan} have been developed to restore clear images from blurry images directly. However, as demonstrated in \cite{dong2020deep}, when the blur kernels are given, these methods do not perform well compared to the non-blind deconvolution methods. 
As non-blind deconvolution is vital for image restoration, we focus on this problem and develop a simple and effective approach to restoring high-quality images.

\begin{figure}[t]
    \centering
    \includegraphics[width=0.9\textwidth, keepaspectratio]{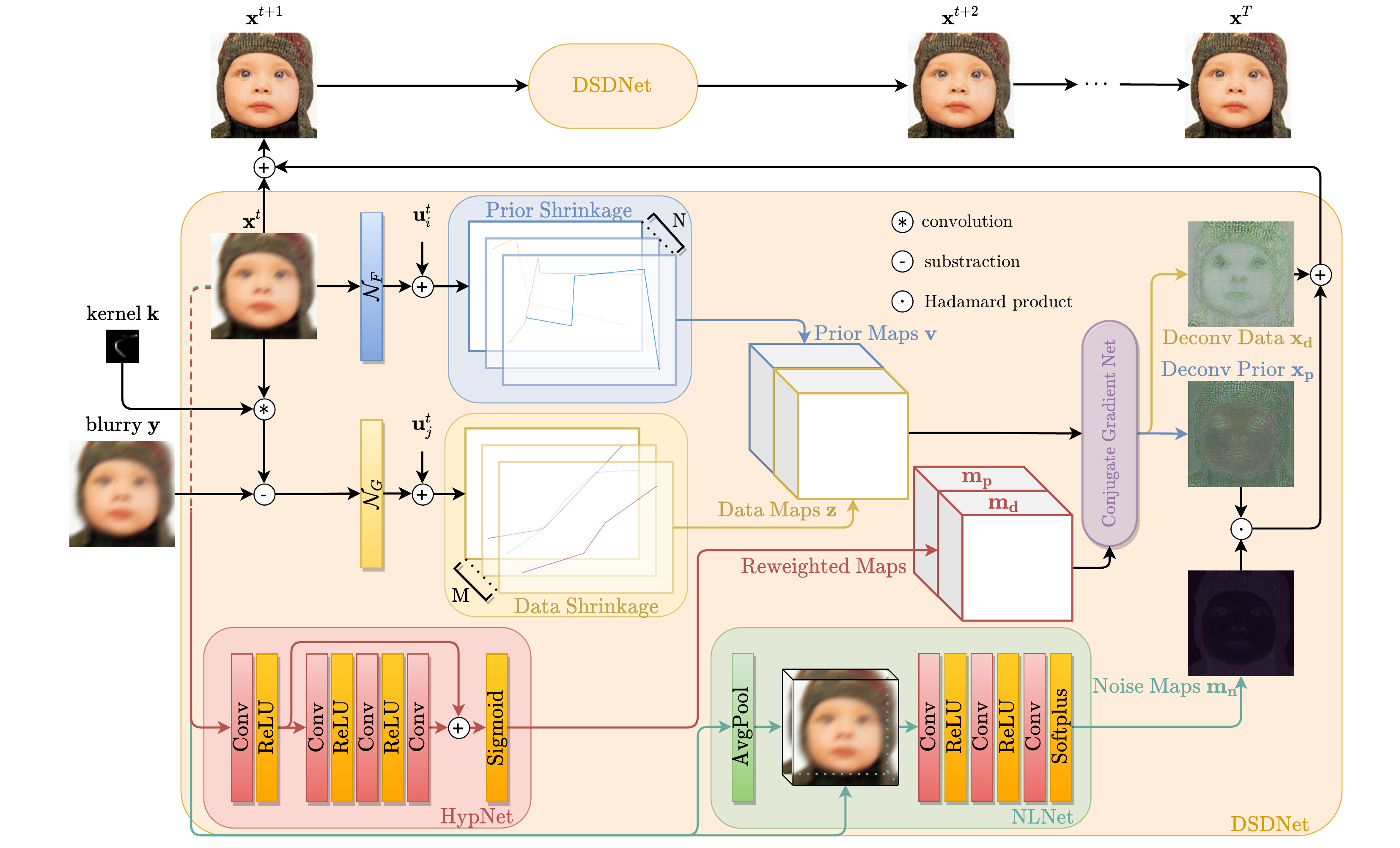}
    \caption{Overview of the proposed method. The blue blocks and lines are the layers and flow of the regularization terms; the yellow ones are that of data terms. The HypNet is responsible for the reweighted maps the NLNet learns to control the weights of regularization and data terms according to the local noise level.} 
    \label{fig:ADMNN}
    
\end{figure}
\begin{figure}[h]
    \centering
    \includegraphics[width=0.7\linewidth, keepaspectratio]{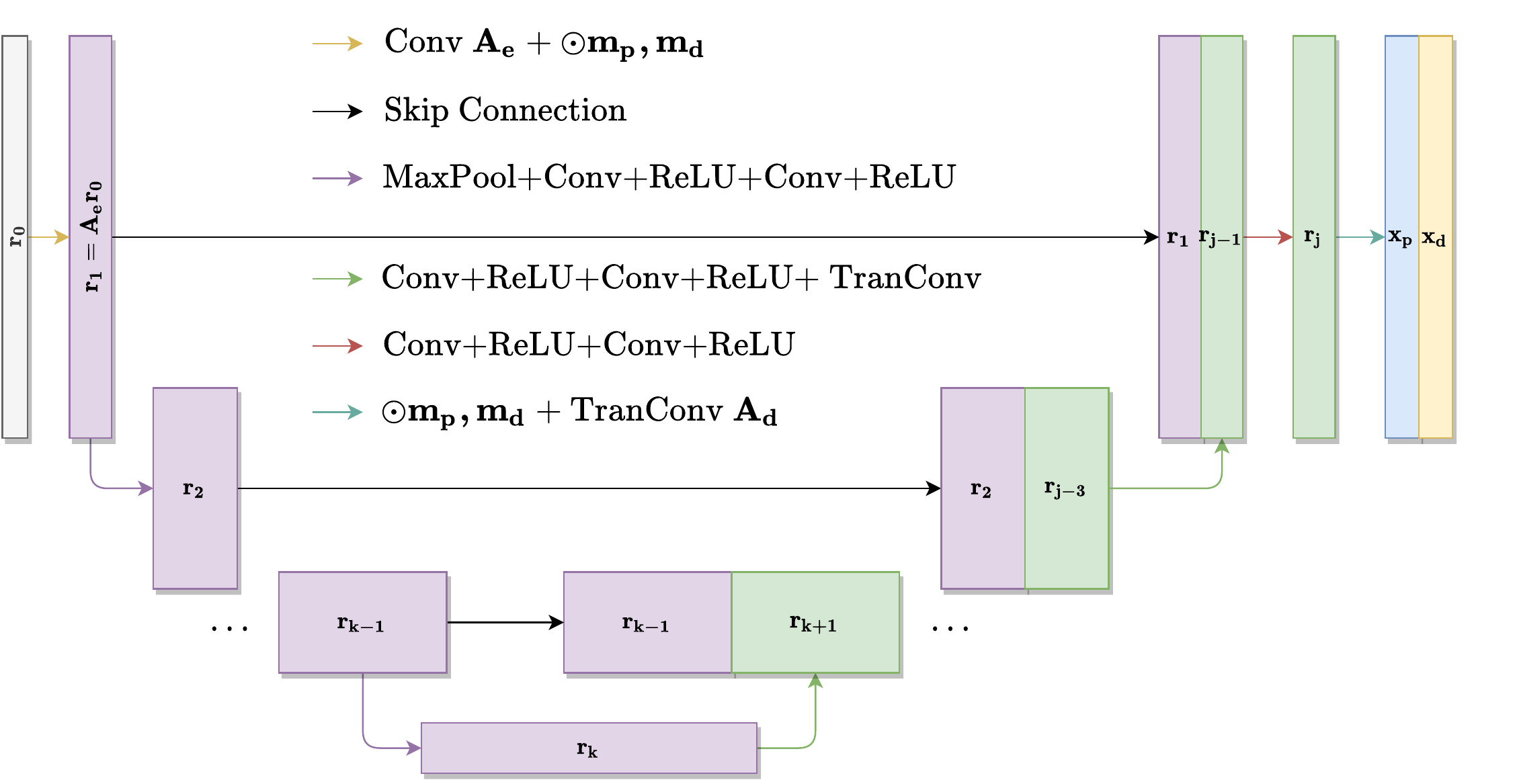}
    
    \caption{Network architecture of the proposed CGNet. The residual vector $\mathbf{r_0}=\mathbf{b-Ax}^t$ is the input feature map,
the operation $\mathbf{A}$, the convolution $\mathbf{A_e}$ and the transposed convolution $\mathbf{A_d}$ are composed of the input filters $\mathbf{F_\mathnormal{i}}$, $\mathbf{G_\mathnormal{j}}$ and $\mathbf{H}$.
The reweighted maps $\mathbf{m_p}$, $\mathbf{m_d}$ are multiplied between $\mathbf{A_e}$ and $\mathbf{A_d}$ as the IRLS.} 
    \label{fig:unet}
    
\end{figure}


\section{Revisiting Deep Unrolling-Based Methods}

\label{sec: revisting}
We first revisit deep unrolling-based methods for image deconvolution to motivate our work.
Mathematically, the degradation process of the image blur is usually formulated as:
\begin{equation}
\label{eq:model}
y = k \ast x + n,
\end{equation}
where $\ast$ denotes the convolution operator; $y$, $k$, $x$ and $n$ denote the blurry image, the blur kernel, the latent image and noise, respectively. 
With the known kernel $k$, we usually use formulate the deconvolution as a maximum-a-posteriori (MAP) problem:
\begin{equation}
\label{eq:map-bayes}
x = \arg\max_{x}p(x|y, k) = \arg\max_{x}p(y|x, k)p(x),
\end{equation}
where $p(y|x, k)$ is the likelihood of the observation (blurry) $y$, while $p(x)$ denotes an image prior of the latent image $x$.
This equation is equivalent to
\begin{equation}
\label{eq:MAP}
\min_{x}R(x)+D(y-k\ast x),
\end{equation}
where $R(x)$ and $D(y-k\ast x)$ denote the regularization term and data term.
In addition, the data term is usually modeled in the form of $\ell_2$-norm, then \eqref{eq:MAP} can be rewritten as
\begin{equation}
\label{eq:obj_MAP}
\min_{\mathbf{x}}\frac{1}{2}\|\mathbf{y-Hx}\|_2^2 + R(\mathbf{x}),
\end{equation}
where $\mathbf{x}$, $\mathbf{y}$ denote the vector forms of $x$ and $y$, respectively; $\mathbf{H}$ denotes the Toeplitz matrix of the blur kernel $k$.
The ADMM method for image deconvolution is usually achieved by solving:
\begin{equation}
\label{eq:hqs-ojb}
\min_{\mathbf{x}, \mathbf{u}}\frac{1}{2}\|\mathbf{y-Hx}\|_2^2 + R(\mathbf{v}) + \mathbf{u}^{\top}(\mathbf{v-x}) 
+\frac{\rho}{2}\|\mathbf{v- x}\|_2^2,
\end{equation}
where $\mathbf{v}$ is an auxiliary variable, $\mathbf{u}$ is a Lagrangian multiplier, $\rho$ is a weight parameter. 

The solution of~\eqref{eq:hqs-ojb} can be obtained by alternatively solving:
\begin{subequations}
\label{eq:deblurring-model-x-modified-u-sol}
\begin{align}
&\mathbf{x}^{t+1} = \min_{\mathbf{x}}\|\mathbf{y-Hx}\|_2^2 + 
\rho\|\mathbf{v}^t-\mathbf{x} + \frac{\mathbf{u}^t}{\rho}\|_2^2, 
\label{eq:x-subproblem}\\
&\mathbf{v}^{t+1} = \min_{\mathbf{v}}\frac{\rho}{2}\|\mathbf{v}-\mathbf{x}^{t+1} + 
\frac{\mathbf{u}^t}{\rho}\|_2^2 +  R(\mathbf{v}), \label{eq:v-subproblem}\\ 
& \mathbf{u}^{t+1} = \mathbf{u}^t + \rho(\mathbf{v}^{t+1}-\mathbf{x}^{t+1}).\label{eq:u-subproblem}
\end{align}
\end{subequations}
Existing methods \cite{fcn/cvpr17,zhang2017learning,zhang2021plug} usually solve~\eqref{eq:x-subproblem} via fast Fourier transform (FFT) or Conjugate Gradient methods. 
For~\eqref{eq:v-subproblem}, its solution can be represented as a proximal operator:
\begin{equation}
\label{eq:u-sub-problem}
\text{\textbf{prox}}_{\lambda R}(\mathbf{x}^{t+1}-\mathbf{u}^{\prime t}) =
\arg\min_{\mathbf{v}}\frac{1}{2}\|\mathbf{v}-\mathbf{x}^{t+1}+\mathbf{u}^{\prime t}\|_2^2 + \lambda R(\mathbf{v}),
\end{equation}
where $\lambda = 1/\rho$. With $\mathbf{u}^{\prime t} =\lambda \mathbf{u}^t$, the multiplier in the \eqref{eq:deblurring-model-x-modified-u-sol} and \eqref{eq:u-sub-problem} can be absorbed \cite{parikh2014proximal}.
As demonstrated in \cite{schmidt2014shrinkage}, \eqref{eq:u-sub-problem} can be approximated by shrinkage functions. 
Existing methods usually use deep CNN model to approximate the solution of~\eqref{eq:u-sub-problem}. However, simply using the convolution operation followed by the fixed activation functions (e.g., ReLU) cannot model the shrinkage functions well as they are far more complex (e.g., non-monotonic) \cite{schmidt2014shrinkage}.
To better approximate the solution of~\eqref{eq:v-subproblem}, we develop a deep CNN model with the Maxout function \cite{goodfellow2013maxout}, which can effectively approximate proximal functions.
In addition, we note that using FFT to solve~\eqref{eq:x-subproblem} does not obtain better results than the CG method demonstrated by \cite{levin2007image,levin2007user}. However, the CG method is time-consuming and unstable in deep networks (see Section~\ref{subsec:ablation} for more detail).
To overcome this problem, we learn a differentiable CG network to restore a clear image more efficiently and effectively. 


\section{Proposed Method}

\label{sec:method}

Different from existing methods that simply learn the regularization term or the data term \cite{fcn/cvpr17,bigdeli2017deep,li2019blind}, we formulate both the data term and the regularization term as the learnable ones:

\begin{equation}
\label{eq:obj_ADMM}
\begin{aligned}
&\min\limits_{\mathbf{u},\mathbf{v},\mathbf{x},\mathbf{z}}\sum\limits_{i=1}^{N}R_i(\mathbf{v}_i)+\sum\limits_{j=1+N}^{M+N}R_j(\mathbf{z}_j)\quad 
\\&s.t.\quad \mathbf{F}_i\mathbf{x}=\mathbf{v}_i, \quad \mathbf{G}_j(\mathbf{y}-\mathbf{Hx})=\mathbf{z}_j,
\end{aligned}
\end{equation}
where $R_i$ denotes the $i$-th learnable function; $\mathbf{v}_i$ and $\mathbf{z}_j$ are auxiliary variables that correspond to the regularization and data terms; $\mathbf{F}_i$ and $\mathbf{G}_j$ are the $i$-th and $j$-th learnable filters for regularization and data, respectively. 

By introducing the Lagrangian multipliers $\mathbf{u}_i$ and $\mathbf{u}_j$ corresponding to the regularization and data terms, we can solve~\eqref{eq:obj_ADMM} using the ADMM method by:
%
%
%
\begin{subequations}
\label{eq:ADMM_all}
\begin{align}
&\mathbf{v}_i^{t+1} = \text{\textbf{prox}}_{\lambda_iR_i}(\mathbf{F}_i\mathbf{x}^t+\mathbf{u}_i^t),
\label{eq:ADMM_v}\\
&\mathbf{z}_j^{t+1} =  \text{\textbf{prox}}_{\lambda_jR_j}(\mathbf{G}_j(\mathbf{y-Hx}^t)+\mathbf{u}_j^t), \label{eq:ADMM_z}\\
&\left(\sum_{i=1}^N\rho_i\mathbf{F}_i^\top \mathbf{F}_i + \sum_{j=N+1}^{N+M}\rho_j\mathbf{H}^\top \mathbf{G}_j^\top \mathbf{G}_j\mathbf{H}\right) \mathbf{x}^{t+1} \nonumber \\
=&\Bigg(\sum_{i=1}^N\rho_i\mathbf{F}_i^\top(\mathbf{v}_i^{t+1}-\mathbf{u}_i^{t})+ 
\sum_{j=N+1}^{N+M}\rho_j\mathbf{H}^\top \mathbf{G}_j^\top(\mathbf{G}_j\mathbf{y}-\mathbf{z}^{t+1}_j+\mathbf{u}_j^{ t}) \Bigg) \label{eq:ADMM_x}\\ 
&\mathbf{u}_i^{t+1} = \mathbf{u}_i^{ t}+\mathbf{F}_i\mathbf{x}^{t+1}-\mathbf{v}_i^{t+1},\label{eq:ADMM_ui} \\
&\mathbf{u}_j^{t+1} = \mathbf{u}_j^{t}+\mathbf{G}_j(\mathbf{y}-\mathbf{H}\mathbf{x}^{t+1})-\mathbf{z}_j^{t+1}.\label{eq:ADMM_uj}  
\end{align}
\end{subequations}

%
%
In the following, we will develop deep CNN models with a Maxout layer to approximate the functions of~\eqref{eq:ADMM_v} and~\eqref{eq:ADMM_z}. Moreover, we design a simple and effective deep CG network to solve \eqref{eq:ADMM_x}.

\subsection{Network Architecture}

\label{subsec:network}
This section describes how to design our deep CNN models to effectively solve\eqref{eq:ADMM_v} -\eqref{eq:ADMM_x} .

\ph{Learning Filters $\mathbf{F}_i$ and $\mathbf{G}_j$.}
To learn filters $\mathbf{F}_i$ and $\mathbf{G}_j$, we develop two networks ($\mathcal{N}_F$ and $\mathcal{N}_G$), each containing one convolutional layer. The convolutional layer of $\mathcal{N}_F$ has $N$ filters of $7\times 7$ pixels, and the convolutional layer of $\mathcal{N}_G$ contains $M$ filters of the same size. The $\mathcal{N}_F$ and $\mathcal{N}_G$ are applied to $\mathbf{x}^t$ and $\mathbf{y-Hx}^t$ to learn the filters $\mathbf{F}_i$ and $\mathbf{G}_j$, respectively.

\ph{Learning Discriminative Shrinkage Functions for \eqref{eq:ADMM_v} and \eqref{eq:ADMM_z}.}
To better learn the unknown discriminative shrinkage functions of \eqref{eq:ADMM_v} and \eqref{eq:ADMM_z}, we take advantage of Maxout layers \cite{goodfellow2013maxout}.
Specifically, the convolutional Maxout layer consists of two Maxout units. 
Each Maxout unit contains one convolutional layer followed by a channel-wise Max-pooling layer.
Given an input feature map $\mathbf{X}\in \mathbb{R}^{H \times W \times C}$ and the output feature map $\mathbf{X^o} \in \mathbb{R}^{H \times W \times KC}$ of the  convolutional layer, a Maxout unit is achieved by:
\begin{equation}
    \label{eq:maxout_unit}
    o_{h,w,c}(\mathbf{X}) = \max_{j \in\left[0, K\right)} x^o_{h,w,c\times K +j},
\end{equation}
where $h\in\left[0,H\right)$, $w\in\left[0,W\right)$ and $c\in\left[0,C\right)$; the $x^o_{h,w,c}$ is the element of $\mathbf{X^o}$ at the position $(h, w, c)$, and the $o_{h,w,c}$ is the function to output the $(h, w, c)$-th element of the output tensor $\mathbf{O}$.
In our implementation, we have $\mathbf{O}\in\mathbb{R}^{H \times W \times C}$ is of the same size as the input $\mathbf{X}$ and $K=4$.

With two Maxout units, we acquire two output features,  $\mathbf{O}_1$ and $\mathbf{O}_2$; the final output tensor of the Maxout layer is their difference,  $\mathbf{O}_1 - \mathbf{O}_2$.
We note that Maxout networks are universal approximators that can effectively approximate functions. Thus, we use it to obtain the solutions of \eqref{eq:ADMM_v} and \eqref{eq:ADMM_z}.
%
%

\ph{Learning a Differentiable CG Network for~\eqref{eq:ADMM_x}.}
As stated in Section~\ref{sec: revisting}, although using FFT with boundary processing operations (e.g., edge taper and Laplacian smoothing~\cite{liu/icip08}) can efficiently solve~\eqref{eq:ADMM_x}, the results are not better than the CG-based solver, which can be observed in Table~\ref{tb:ablation}. However, using a CG-based solver is time-consuming. To generate latent clear images better, we develop a differentiable CG network to solve~\eqref{eq:ADMM_x}.
The CG method is used to solve the linear equation:
\begin{equation}
\label{eq:linear-model}
\mathbf{Ax}^{t+1}=\mathbf{b},
\end{equation}
where $\mathbf{A}$ corresponds to the first term in \eqref{eq:ADMM_x} and $\mathbf{b}$ to the last term in \eqref{eq:ADMM_x}.
Given $\mathbf{x}\in\mathbb{R}^d$, the CG method recursively computes conjugate vectors $\mathbf{p}_l$ and find the difference between desired $\mathbf{x}^{t+1}$ and initial input $\mathbf{x}^t$ as $$\mathbf{x}^{t+1}-\mathbf{x}^t=\mathbf{s_L}=\sum_{l=0}^L \alpha_l\mathbf{p}_l,$$
where $L$ is the iteration number upper-bounded by $d$, and $\alpha_l$ is the weight calculated with $\mathbf{p}_l$.

However, if the size of matrix $\mathbf{A}$ is large, using CG to solve~\eqref{eq:linear-model} needs high computational costs. 
To overcome this problem, we develop a differentiable CG network based on a U-Net to compute the $\mathbf{s_L}$. The network design is motivated by the following reasons:
\begin{compactitem}
\item As one of Krylov subspace methods \cite{barrett1994templates}, the solution $\mathbf{s_L}$ can be found in the Krylov subspace $\mathcal{K}_L(\mathbf{A,r_0})=\text{span}\{\mathbf{r_0, Ar_0, \dots, A^{\mathnormal {L}-\mathrm{1}}r_0}\}$, where $\mathbf{r_0}=\mathbf{b-Ax}^t$ is the residual vector.
In other words, the CG method is a function of $\mathbf{A}$ and $\mathbf{r_0}$.
Our CGNet takes $\mathbf{r_0}$ as input and $\mathbf{A}$ as parts of the network. Its output is $\mathbf{s_L}$, which behaves as the CG method.
\item For a typical deconvolution problem, $\mathbf{A}$ is composed of convolution $\mathbf{A_e}$ and transpose convolution $\mathbf{A_d}$ pairs as the first term in \eqref{eq:ADMM_x}.
$\mathbf{A_e}$ stands for the operation of $\mathbf{G_\mathnormal{j}H}$ and $\mathbf{F_\mathnormal{i}}$, and $\mathbf{A_d}$ for $\mathbf{H^\top G^\top_\mathnormal{j}}$ and $\mathbf{F^\top_\mathnormal{i}}$.
This observation can be intuitively connected to an encoder-decoder architecture, so we integrate $\mathbf{A_e}$ into the encoder and $\mathbf{A_d}$ into the decoder.
\item The Conjugate Gradient method is sensitive to noise \cite{liu2012modifications,ryabtsev2020error}. With an encoder-decoder architecture, U-Net is robust to noise.
\item As the Conjugate Gradient method is a recursive algorithm, U-Net computes feature maps  in a recursive fashion.
\end{compactitem}

In the practical CG iterations, $\mathbf{A_e}$ and $\mathbf{A_d}$ are usually updated with iterative reweighted least squares (IRLS) to utilize the sparsity of priors \cite{levin2007user}. 
We design a simple HypNet to estimate these weights. The network architecture of the HypNet is shown in Fig.~\ref{fig:ADMNN}.
In addition, we note that the values of $\rho_i$ and $\rho_j$ in \eqref{eq:ADMM_x} depend on the noise level. 
We design a simple NLNet to estimate the noise map $\mathbf{m_n}$, which plays a role similar to $\rho_i$ and $\rho_j$ (see Fig.~\ref{fig:ADMNN}). 
In contrast to most conventional methods, the NLNet computes the weight for each pixel, which is locally adaptive.
%

\section{Experimental Results}

\label{sec:result}
\subsection{Datasets and Implementation Details} 

\label{subsec:detail}
\ph{Training Dataset.} 
Similar to  \cite{dong2020deep,dong2021learning}, the training data is composed of 4,744 images from the Waterloo Exploration dataset \cite{ma2016waterloo} and 400 images from the Berkeley segmentation dataset (\textsc{BSD}) \cite{martin2001database}. 
To synthesize blurry images, we first generate 33,333 blur kernels by \cite{schmidt2015cascades}, where the sizes of these generated blur kernels range from $13\times 13$ pixels to $35\times 35$ pixels.
We first crop image patches of $128\times128$ pixels from each image, and then we randomly use generated blur kernels to generate blurry images. Each blurry image is randomly added Gaussian noise with noise levels from 1\% to 5\%. 

\ph{Test Datasets.} We evaluate our method on both synthetic datasets and real ones. 
For the synthetic case, we use the 100 images from \textsc{BSD100} and 100 kernels  by \cite{schmidt2015cascades} to generate blurry images, similar to training data.  
We also use the \textsc{Set5} \cite{bevilacqua2012low} dataset with the kernels generated by  \cite{chakrabarti2016neural} as our test dataset. 
In addition, we use the datasets by \textsc{Levin} \cite{levin2009understanding} and \textsc{Lai} \cite{lai2016comparative} for evaluation.

For the real-world dataset, we evaluate our model on the data of \textsc{Pan} \cite{pan2016blind}, where 23 blurry images and 23 kernels estimated by their method are contained.
%
\begin{table}[h]
\footnotesize
    \caption{Configuration of four models. $T$ is the number of stages, i.e., how many duplicates are in the whole model. $M$ and $N$ denote the number of filters $\mathbf{F}_i$and $\mathbf{G}_i$, respectively. 
   The PSNR (dB) and execution time are tested on \textsc{Set5}.
    }
    \centering
    \begin{tabular}{l c c c c}
    \toprule
         &Feather &Light &Heavy &Full \\ \midrule
         $T$ &2 &3 &3 &4 \\
         $M,N$ &24 &24 &49 &49 \\  \midrule
         PSNR &32.51 &32.78 &33.11 &33.43\\
         Second  &1.893 &2.191 &2.672 &3.065\\
         \bottomrule
    \end{tabular}
    \label{tb:configure}
    
\end{table}

\ph{Implementation Details.} We train the networks using the ADAM \cite{kingma2014adam} optimizer with default parameter settings. The batch size is 8. The total training iterations is 1 million, and the learning rate is from $1\times10^{-4}$ to $1\times 10^{-7}$. 
We gradually decay the learning rate to $1\times 10^{-7}$ every 250,000 iterations and reset it to $1\times10^{-4}$, $5\times10^{-5}$ and $2.5\times10^{-5}$ at the iteration of 250,001, 500,001 and 750,001, respectively.
To constrain the network training, we apply the commonly used $\ell_1$-norm loss to the ground truth and the network output $\mathbf{x}^T$.
The data augmentation (including $\pm$90\textdegree and 180\textdegree rotations, vertical and horizontal flipping) is used. 
In this paper, we train 4 models of different sizes, i.e., Feather, Light, Heavy and Full, whose configurations and simple results on \textsc{Set5} are shown in Table~\ref{tb:configure}.
It is worth noting that all the stages contain their own parameters and are end-to-end trained rather than share weights \cite{zhang2020deep} or progressively trained \cite{fcn/cvpr17}.
%
%
We implement the networks in Pytorch \cite{paszke2019pytorch} and train on one NVIDIA RTX 3090 GPU. 

\subsection{Quantitative Evaluation} 

\label{subsec:quantitative}


We compare the proposed DSDNet with 7 state-of-the-art methods including IRCNN \cite{zhang2017learning}, SARFL \cite{ren2019simultaneous}, ADM\_UDM \cite{ko2020deep}, CPCR \cite{eboli2020end}, KerUNC \cite{nan2020deep}, VEM \cite{nan2020variational}, DWDN \cite{dong2020deep}, SVMAP \cite{dong2018learning} and DRUNet \cite{zhang2021plug}. 
%
These methods are fine-tuned using the same training dataset as Section~\ref{subsec:detail} and choose the best models from fine-tuned and the original ones for comparison.

PSNR and SSIM \cite{wang2004image} are used for quantitative evaluation. All the quantitative evaluations are conducted without border cropping for fair comparisons.

Table~\ref{tb:Levin} shows quantitative evaluation results on the synthetic datasets. 
The proposed method generates results with higher PSNR and SSIM values.
In addition, we note that the proposed light-weighted model generates favorable results against the state-of-the-art, showing the effectiveness of the proposed algorithm. 
Due to the space limit, we only present the evaluation results of the Full DSDNet hereafter.

\begin{table}[t]
\caption{Average PSNR(dB)/SSIM of the deblurring results with Gaussian noise using different methods. We highlight the \fst{best} and the \snd{second} best results. Our Full DSDNet wins first place, while our Light one also performs favorably against these state-of-the-art methods.}
\label{tb:Levin}
\centering
\tabcolsep=1pt
\scriptsize
\resizebox{\textwidth}{!}{
\begin{tabular}{c c c c c c c c c c c c c}
\toprule
\mr{2}{Dataset} &\mr{2}{noise} &IRCNN \cite{zhang2017learning} &SFARL\cite{ren2019simultaneous} &ADM\_UDM \cite{ko2020deep} &CPCR \cite{eboli2020end} &KerUNC \cite{nan2020deep} &VEM \cite{nan2020variational} &DWDN \cite{dong2020deep} &SVMAP \cite{dong2021learning} &DRUNet \cite{zhang2021plug} &DSDNet(Light) &DSDNet(Full)\\
& &PSNR / SSIM &PSNR / SSIM &PSNR / SSIM &PSNR / SSIM &PSNR / SSIM &PSNR / SSIM &PSNR / SSIM &PSNR / SSIM &PSNR / SSIM &PSNR / SSIM &PSNR / SSIM\\
\midrule
\mr{3}{\shortstack{\textsc{Levin}\\\cite{levin2009understanding}}}
&$1\%$ &30.61 / 0.883 &25.41 / 0.600 &31.48 / 0.922   &28.43 / 0.858 &32.02 / 0.928 &32.05 / 0.927 &34.89 / 0.957 &35.24 / \snd{0.962} &31.94 / 0.922 &\snd{35.48} / 0.960 &\fst{36.62} / \fst{0.965}\\  
&$3\%$ &29.70 / 0.864 &16.82 / 0.255 &28.61 / 0.812   &25.61 / 0.765 &21.72 / 0.416 &29.47 / 0.867 &31.94 / 0.916 &31.20 / 0.893 &30.86 / 0.905 &\snd{32.13} / \snd{0.918} &\fst{32.89} / \fst{0.925}\\ 
&$5\%$ &28.98 / 0.854 &13.07 / 0.157 &27.83 / 0.827   &23.68 / 0.703 &18.25 / 0.272 &27.79 / 0.819 &30.21 / 0.883 &30.12 / 0.876 &29.79 / 0.880 &\snd{30.24} / \snd{0.883} &\fst{30.94} / \fst{0.893}\\ \midrule                                                                                                                               
\mr{3}{\shortstack{\textsc{BSD100}\\\cite{martin2001database}}}                                                                 
&$1\%$ &29.20 / 0.817 &24.21 / 0.568 &29.39 / 0.836   &28.77 / 0.829 &29.23 / 0.829 &29.54 / 0.848 &31.10 / 0.881 &\snd{31.52} / 0.888 &30.36 / 0.872 &31.50 / \snd{0.892} &\fst{32.01} / \fst{0.898}\\ 
&$3\%$ &27.54 / 0.762 &15.80 / 0.245 &26.92 / 0.722   &25.96 / 0.712 &22.10 / 0.430 &27.09 / 0.746 &28.47 / 0.797 &27.94 / 0.762 &28.10 / 0.798 &\snd{28.73} / \snd{0.812} &\fst{29.08} / \fst{0.820}\\
&$5\%$ &27.04 / 0.756 &12.56 / 0.146 &26.04 / 0.697   &25.75 / 0.688 &18.99 / 0.297 &26.11 / 0.698 &27.50 / 0.762 &27.59 / 0.763 &27.19 / 0.767 &\snd{27.64} / \snd{0.774} &\fst{27.96} / \fst{0.782}\\ \midrule  
\mr{3}{\shortstack{\textsc{Set5}\\\cite{bevilacqua2012low}}}                                                                                          
&$1\%$ &30.15 / 0.853 &26.21 / 0.632 &30.52 / 0.868   &30.59 / 0.875 &30.45 / 0.864 &31.00 / 0.875 &32.18 / 0.893 &32.31 / 0.892 &30.84 / 0.881 &\snd{32.78} / \snd{0.899} &\fst{33.43} / \fst{0.905}\\   
&$3\%$ &28.66 / 0.813 &15.50 / 0.211 &27.64 / 0.709   &27.94 / 0.799 &21.39 / 0.376 &28.40 / 0.804 &29.54 / 0.838 &28.78 / 0.812 &29.21 / 0.841 &\snd{29.94} / \snd{0.843} &\fst{30.40} / \fst{0.851}\\ 
&$5\%$ &27.55 /	0.789 &11.91 / 0.122 &26.75 / 0.756   &26.64 / 0.754 &17.74 / 0.241 &26.46 / 0.732 &28.13 / \snd{0.806} &28.02 / 0.793 &27.85 / 0.805 &\snd{28.46} / 0.804 &\fst{28.89} / \fst{0.814}\\
\bottomrule
\end{tabular}}
\medskip
\caption{Evaluation on the dataset \textsc{Lai} \cite{lai2016comparative}. The best and second best results are highlighted as Table~\ref{tb:Levin}. The \textsc{Saturation} results of SVMAP \cite{dong2021learning} are obtained by the model specifically trained for saturation scenes.}
\label{tb:Jason}
\centering
\tabcolsep=3pt
\footnotesize
\resizebox{\textwidth}{!}{
\begin{tabular}{l c c c c c c c c }
\toprule
\mr{2}{Subset}   &IRCNN \cite{zhang2017learning}  &ADM\_UDM \cite{ko2020deep} &KerUNC \cite{nan2020deep} &VEM \cite{nan2020variational} &DWDN \cite{dong2020deep} &SVMAP \cite{dong2021learning} &DRUNet \cite{zhang2021plug} &DSDNet\\
&PSNR / SSIM &PSNR / SSIM &PSNR / SSIM &PSNR / SSIM &PSNR / SSIM &PSNR / SSIM &PSNR / SSIM &PSNR / SSIM\\
\midrule
\textsc{Manmade}    &20.47 / 0.604 &22.43 / 0.724 &22.19 / 0.725 &22.71 / 0.780 &\snd{24.02} / \snd{0.836} &23.75 / 0.776 &20.62 /0.613 &\fst{25.44} / \fst{0.859}\\
\textsc{Natural}    &23.26 / 0.636 &25.04 / 0.733 &25.42 / 0.757 &25.29 / 0.752 &25.91 / \snd{0.814} &\snd{26.23} / 0.778 &23.25 /0.630 &\fst{27.01} / \fst{0.837}\\
\textsc{People}     &28.04 / 0.843 &28.81 / 0.866 &28.80 / 0.848 &27.19 / 0.723 &30.02 / \snd{0.905} &\snd{30.88} / 0.899 &28.04 /0.838 &\fst{30.95} / \fst{0.908}\\
\textsc{Saturation} &16.99 / 0.642 &17.57 / 0.627 &17.70 / 0.640 &17.65 / 0.600 &17.90 / 0.695 &\fst{18.75} / \snd{0.733} &17.14 /0.658 &\snd{18.38} / \fst{0.734}\\
\textsc{Text}       &21.37 / 0.828 &25.13 / 0.883 &23.32 / 0.855 &24.92 / 0.853 &25.40 / 0.877 &\snd{25.60} / \snd{0.894} &21.79 /0.829 &\fst{28.13} / \fst{0.920}\\ \midrule
Overall             &22.03 / 0.710 &23.80 / 0.767 &23.49 / 0.765 &23.55 / 0.742 &24.65 / \snd{0.825} &\snd{25.04} / 0.816 &22.17 /0.714 &\fst{25.98} / \fst{0.852}\\
\bottomrule
\end{tabular}}
\end{table}

We then evaluate our method on the \textsc{Lai} \cite{lai2016comparative} dataset. 
Because it contains \textsc{Manmade}, \textsc{Natural}, \textsc{People}, \textsc{Saturation}, and \textsc{Text} subsets, we present the evaluation accordingly.
Table~\ref{tb:Jason} shows that our method performs better than the evaluated methods. 
Similar to Table~\ref{tb:Levin}, our method also achieves the highest PSNR and SSIM in most tests, except for the \textsc{Saturation} subset.
We note that Dong et al. \cite{dong2021learning} specifically train a model for saturation scenes; thus,  this method performs slightly better.
However, our model is only trained with common scenes but comparable in terms of PSNR on the \textsc{Saturation} images. Moreover, the SSIM values of our method are better than  SVMAP \cite{dong2021learning}, demonstrating the efficiency and robustness of our approach.


We also quantitatively evaluate our method on real-world blurry images and estimated kernels from \textsc{Pan} \cite{pan2016blind}.
Since the ground truth images are unavailable, we use the no-reference BRISQUE \cite{mittal2012no} and PIQE \cite{wang2008new} metrics for evaluation. 
Our model achieves the best score in BRISQUE and second place in PIQE, as shown in Table~\ref{tab:NQ}.
As BRISQUE is a metric based on subject scoring, Table~\ref{tab:NQ} shows that our model generates more subjectively satisfying results than other state-of-the-art methods.

\begin{table}[t]
\caption{Quantitative evaluation of real cases \textsc{Pan} \cite{pan2016blind}. 
Non-reference image quality metrics BRISQUE \cite{mittal2012no} and PIQE \cite{venkatanath2015blind} are used.}
\label{tab:NQ}
\centering
\footnotesize
\tabcolsep=3pt
\resizebox{\textwidth}{!}{
\begin{tabular}{l c c c c c c c c}
\toprule
&IRCNN \cite{zhang2017learning}  &ADM\_UDM \cite{ko2020deep} &KerUNC \cite{nan2020deep} &VEM \cite{nan2020variational} &DWDN \cite{dong2020deep} &SVMAP \cite{dong2021learning} &DRUNet \cite{zhang2021plug} &DSDNet \\
\midrule
BRISQUE &43.484 &36.598 &37.816 &\snd{33.663} &34.027 &35.508 &46.774 &\fst{33.129}\\
PIQE    &78.700 &67.605 &65.674 &\fst{44.942} &51.348 &56.032 &81.074 &\snd{49.788}\\
\bottomrule
\end{tabular}}
\end{table}

\begin{figure}[h]
\raggedright
\begin{minipage}[b]{.95\linewidth}
  \subfloat[Blurry input]{
    \begin{minipage}{0.195\linewidth} 
      \centering
      \includegraphics[width=\linewidth]{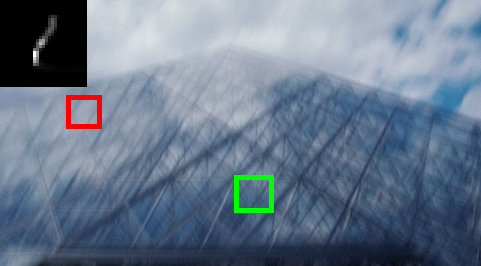}
      \rb{\includegraphics[width=0.499\linewidth, height=0.499\linewidth]{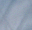}}\hfill
  	  \gb{\includegraphics[width=0.499\linewidth, height=0.499\linewidth]{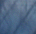}}\hfill
    \end{minipage}
	}
  \subfloat[IRCNN \cite{zhang2017learning}]{
    \begin{minipage}{0.195\linewidth} 
      \centering
      \includegraphics[width=\linewidth]{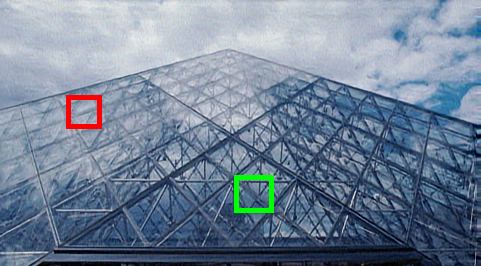}
      \rb{\includegraphics[width=0.499\linewidth, height=0.499\linewidth]{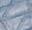}}\hfill
  	  \gb{\includegraphics[width=0.499\linewidth, height=0.499\linewidth]{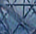}}\hfill
    \end{minipage}
	}
  \subfloat[ADM\_UDM \cite{ko2020deep}]{
    \begin{minipage}{0.195\linewidth} 
      \centering
      \includegraphics[width=\linewidth]{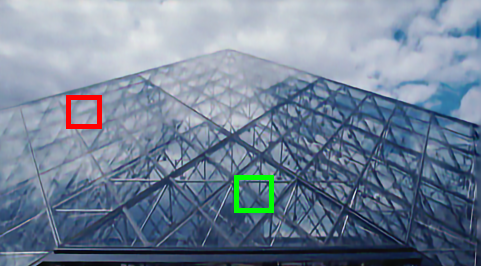}
      \rb{\includegraphics[width=0.499\linewidth,height=0.499\linewidth]{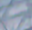}}\hfill
  	  \gb{\includegraphics[width=0.499\linewidth, height=0.499\linewidth]{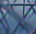}}\hfill
    \end{minipage}
	}
  \subfloat[KerUNC \cite{nan2020deep}]{
    \begin{minipage}{0.195\linewidth} 
      \centering
      \includegraphics[width=\linewidth]{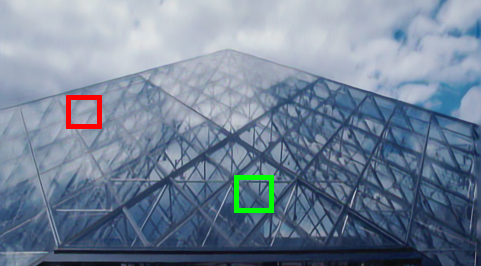}
      \rb{\includegraphics[width=0.499\linewidth, height=0.499\linewidth]{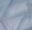}}\hfill
  	  \gb{\includegraphics[width=0.499\linewidth,height=0.499\linewidth]{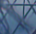}}\hfill
    \end{minipage}
	}
  \subfloat[VEM \cite{nan2020variational}]{
    \begin{minipage}{0.195\linewidth} 
      \centering
      \includegraphics[width=\linewidth]{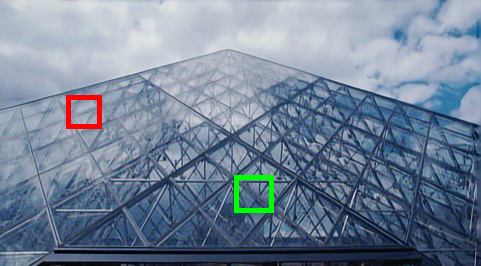}
      \rb{\includegraphics[width=0.499\linewidth, height=0.499\linewidth]{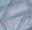}}\hfill
  	  \gb{\includegraphics[width=0.499\linewidth, height=0.499\linewidth]{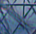}}\hfill
    \end{minipage}
	} \\
  \subfloat[DWDN \cite{dong2020deep}]{
    \begin{minipage}{0.195\linewidth} 
      \centering
      \includegraphics[width=\linewidth]{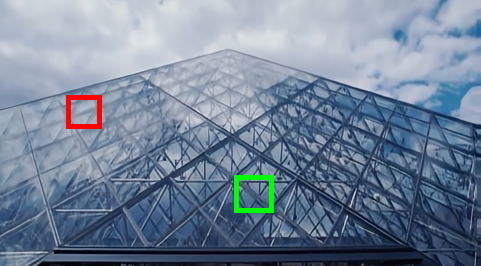}
      \rb{\includegraphics[width=0.499\linewidth, height=0.499\linewidth]{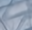}}\hfill
  	  \gb{\includegraphics[width=0.499\linewidth, height=0.499\linewidth]{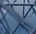}}\hfill
    \end{minipage}
	}
  \subfloat[SVMAP \cite{dong2021learning}]{
    \begin{minipage}{0.195\linewidth} 
      \centering
      \includegraphics[width=\linewidth]{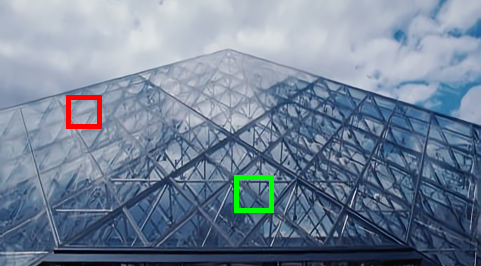}
      \rb{\includegraphics[width=0.499\linewidth, height=0.499\linewidth]{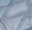}}\hfill
  	  \gb{\includegraphics[width=0.499\linewidth,height=0.499\linewidth]{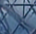}}\hfill
    \end{minipage}
	}
  \subfloat[DRUNet \cite{zhang2021plug}]{
    \begin{minipage}{0.195\linewidth} 
      \centering
      \includegraphics[width=\linewidth]{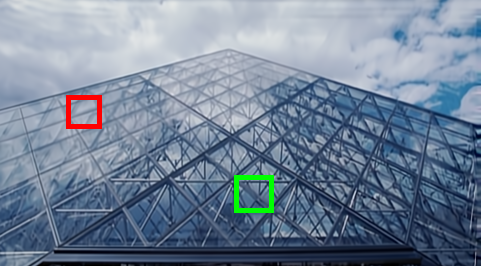}
      \rb{\includegraphics[width=0.499\linewidth, height=0.499\linewidth]{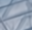}}\hfill
  	  \gb{\includegraphics[width=0.499\linewidth, height=0.499\linewidth]{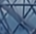}}\hfill
    \end{minipage}
	}
  \subfloat[DSDNet (ours)]{
    \begin{minipage}{0.195\linewidth} 
      \centering
      \includegraphics[width=\linewidth]{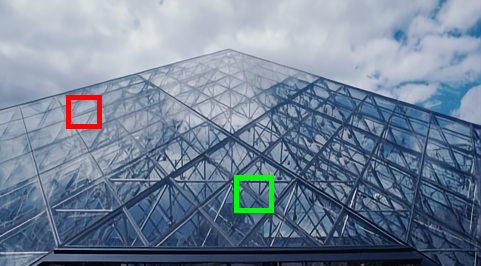}
      \rb{\includegraphics[width=0.499\linewidth, height=0.499\linewidth]{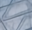}}\hfill
  	  \gb{\includegraphics[width=0.499\linewidth, height=0.499\linewidth]{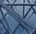}}\hfill
    \end{minipage}
	}
  \subfloat[Ground truth]{
    \begin{minipage}{0.195\linewidth} 
      \centering
      \includegraphics[width=\linewidth]{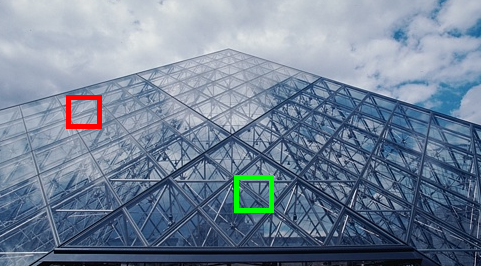}
      \rb{\includegraphics[width=0.499\linewidth, height=0.499\linewidth]{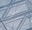}}\hfill
  	  \gb{\includegraphics[width=0.499\linewidth, height=0.499\linewidth]{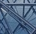}}\hfill
    \end{minipage}
	}\hfill				
  \end{minipage}
	\vfill
      \caption{A synthetic case with $1\%$ Gaussian noise from BSD100 \cite{martin2001database}. Our approach restores the image with finer detailed structures.}
\label{fig:ev1}
\end{figure}

\subsection{Qualitative Evaluation}
\label{subsec:qualitative}
\begin{figure*}[ht]
  \raggedright
  \tiny
  \begin{minipage}[b]{.95\linewidth}
  \subfloat[Blurry input]{
    \begin{minipage}{0.195\linewidth} 
      \centering
      \includegraphics[width=\linewidth]{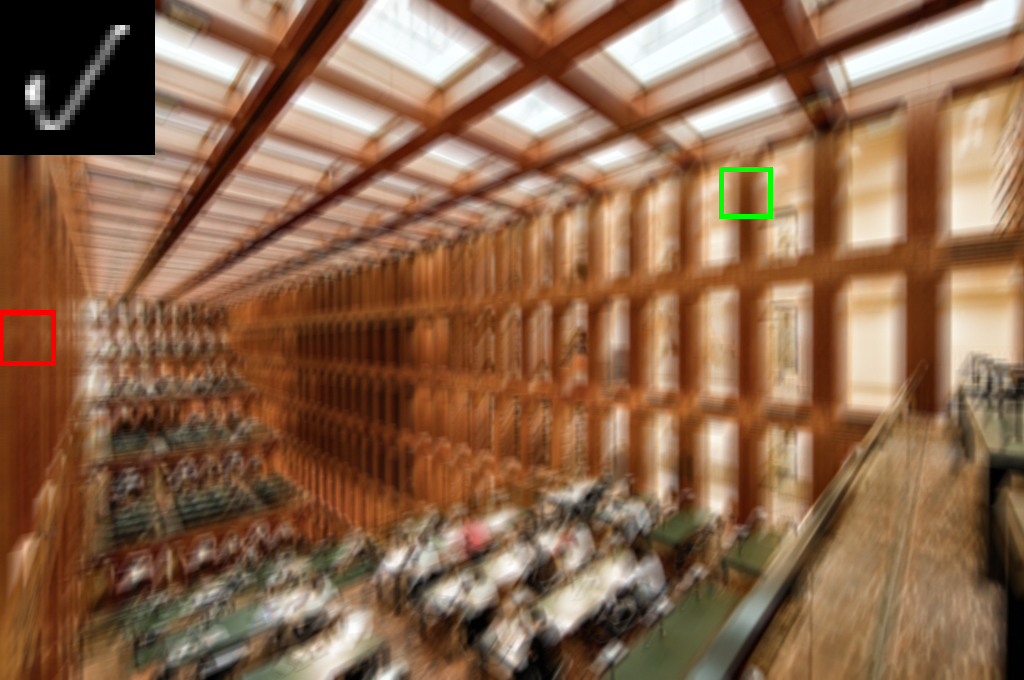}
      \rb{\includegraphics[width=0.499\linewidth,height=0.499\linewidth]{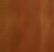}}\hfill
  	  \gb{\includegraphics[width=0.499\linewidth,height=0.499\linewidth]{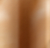}}\hfill
    \end{minipage}
	}
  \subfloat[IRCNN \cite{zhang2017learning}]{
    \begin{minipage}{0.195\linewidth} 
      \centering
      \includegraphics[width=\linewidth]{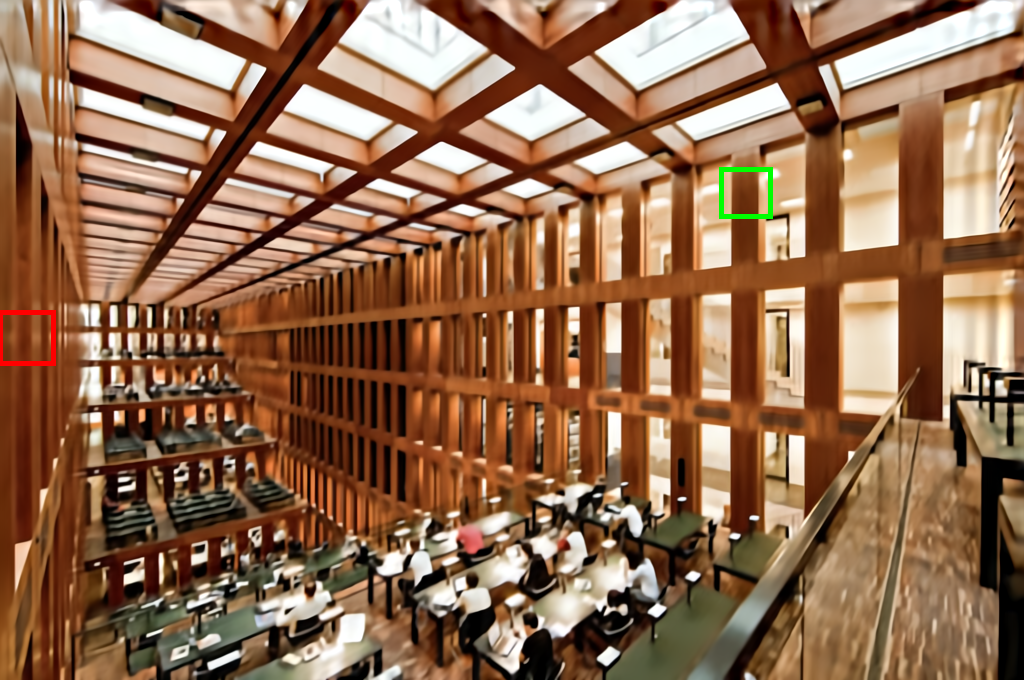}
      \rb{\includegraphics[width=0.499\linewidth, height=0.499\linewidth]{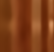}}\hfill
  	  \gb{\includegraphics[width=0.499\linewidth, height=0.499\linewidth]{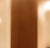}}\hfill
    \end{minipage}
	}
  \subfloat[ADM\_UDM \cite{ko2020deep}]{
    \begin{minipage}{0.195\linewidth} 
      \centering
      \includegraphics[width=\linewidth]{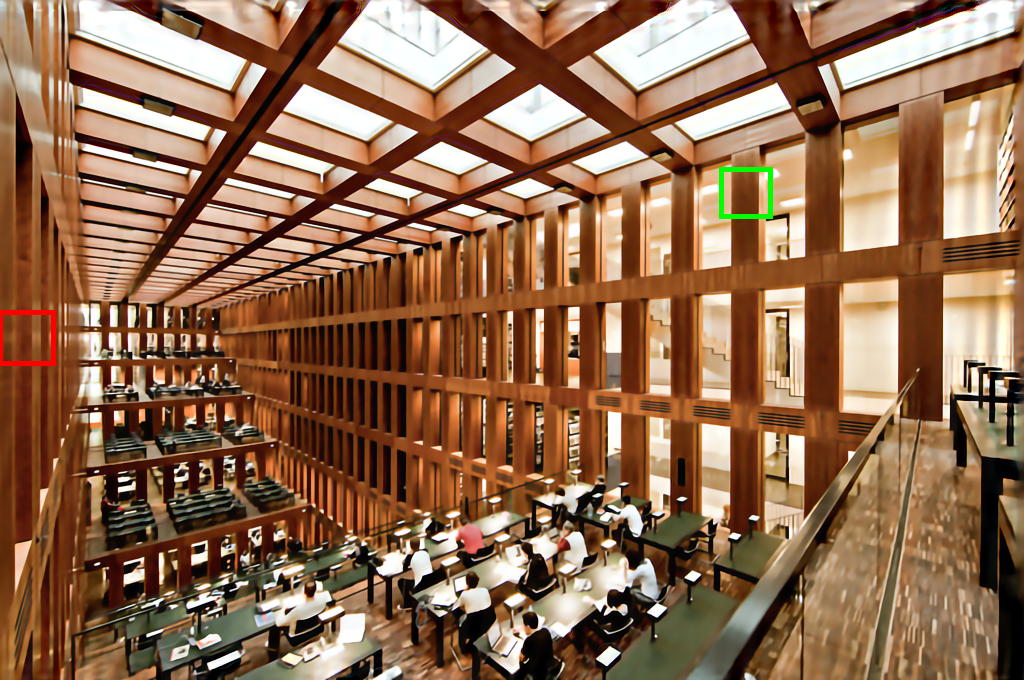}
      \rb{\includegraphics[width=0.499\linewidth, height=0.499\linewidth]{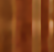}}\hfill
  	  \gb{\includegraphics[width=0.499\linewidth, height=0.499\linewidth]{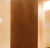}}\hfill
    \end{minipage}
	}
  \subfloat[KerUNC \cite{nan2020deep}]{
    \begin{minipage}{0.195\linewidth} 
      \centering
      \includegraphics[width=\linewidth]{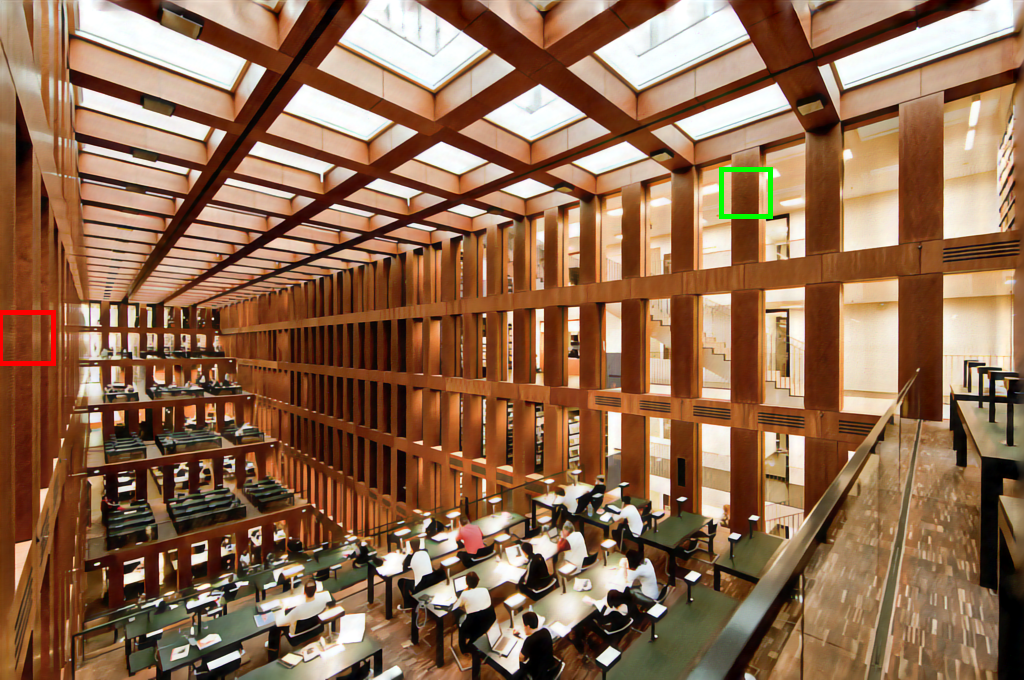}
      \rb{\includegraphics[width=0.499\linewidth, height=0.499\linewidth]{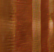}}\hfill
  	  \gb{\includegraphics[width=0.499\linewidth, height=0.499\linewidth]{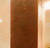}}\hfill
    \end{minipage}
	}
  \subfloat[VEM \cite{nan2020variational}]{
    \begin{minipage}{0.195\linewidth} 
      \centering
      \includegraphics[width=\linewidth]{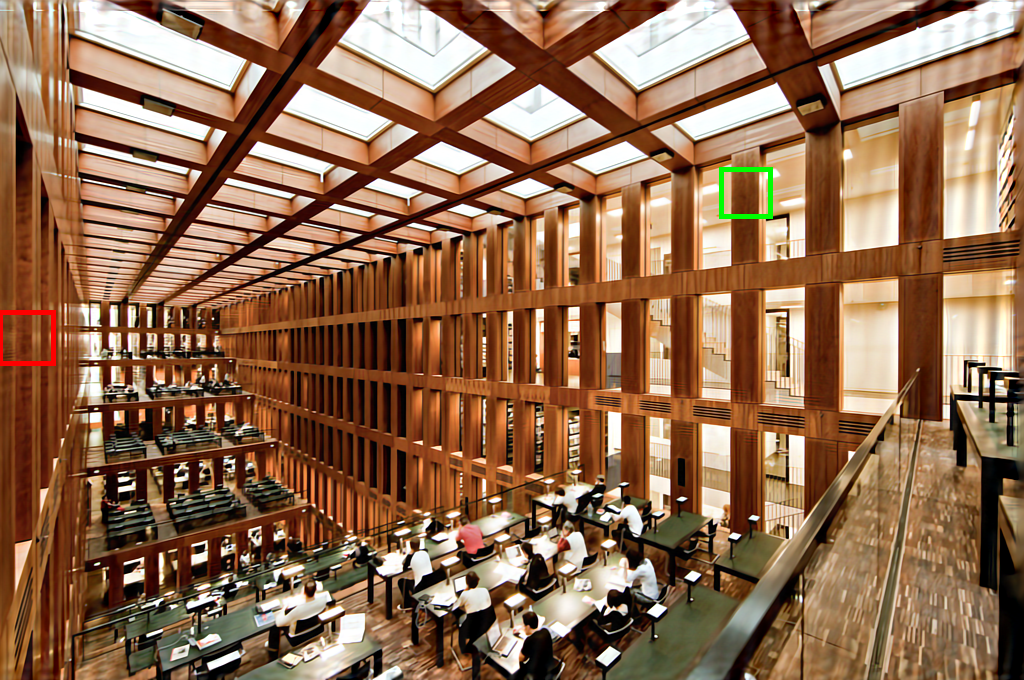}
      \rb{\includegraphics[width=0.499\linewidth, height=0.499\linewidth]{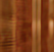}}\hfill
  	  \gb{\includegraphics[width=0.499\linewidth, height=0.499\linewidth]{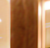}}\hfill
    \end{minipage}
	} \\
  \subfloat[DWDN \cite{dong2020deep}]{
    \begin{minipage}{0.195\linewidth} 
      \centering
      \includegraphics[width=\linewidth]{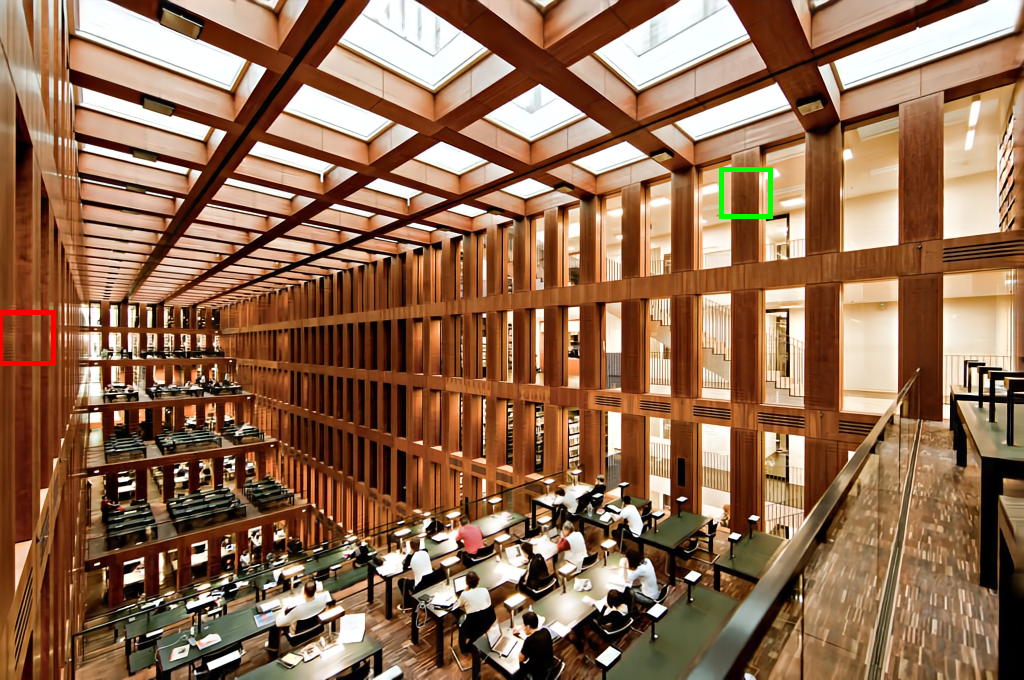}
      \rb{\includegraphics[width=0.499\linewidth, height=0.499\linewidth]{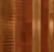}}\hfill
  	  \gb{\includegraphics[width=0.499\linewidth, height=0.499\linewidth]{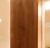}}\hfill
    \end{minipage}
	}
  \subfloat[SVMAP \cite{dong2021learning}]{
    \begin{minipage}{0.195\linewidth} 
      \centering
      \includegraphics[width=\linewidth]{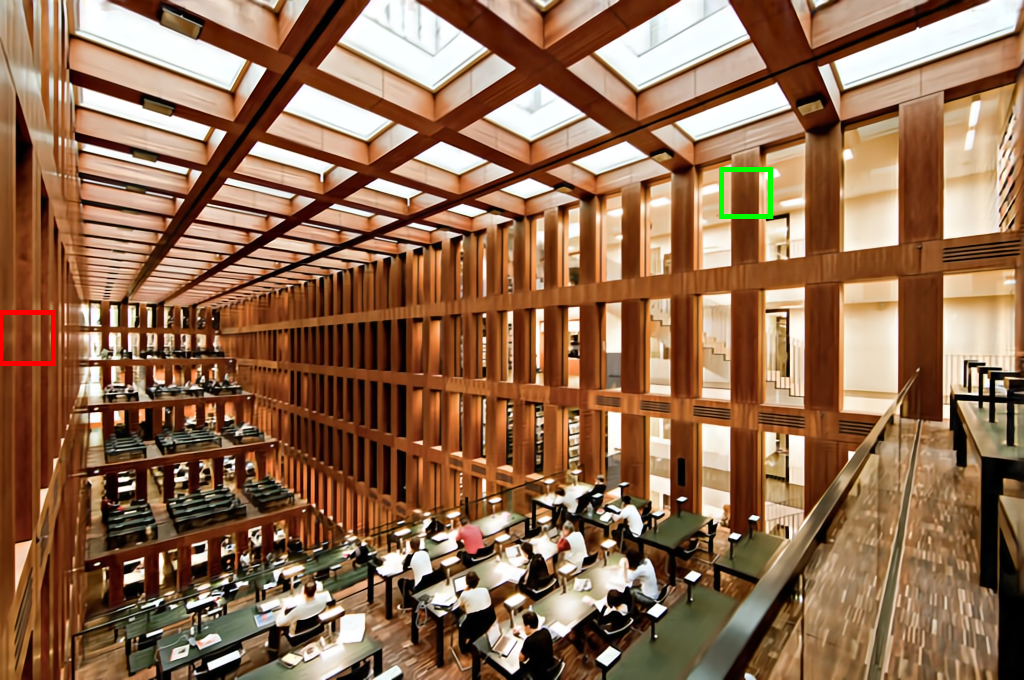}
      \rb{\includegraphics[width=0.499\linewidth, height=0.499\linewidth]{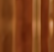}}\hfill
  	  \gb{\includegraphics[width=0.499\linewidth, height=0.499\linewidth]{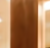}}\hfill
    \end{minipage}
	}
  \subfloat[DRUNet \cite{zhang2021plug}]{
    \begin{minipage}{0.195\linewidth} 
      \centering
      \includegraphics[width=\linewidth]{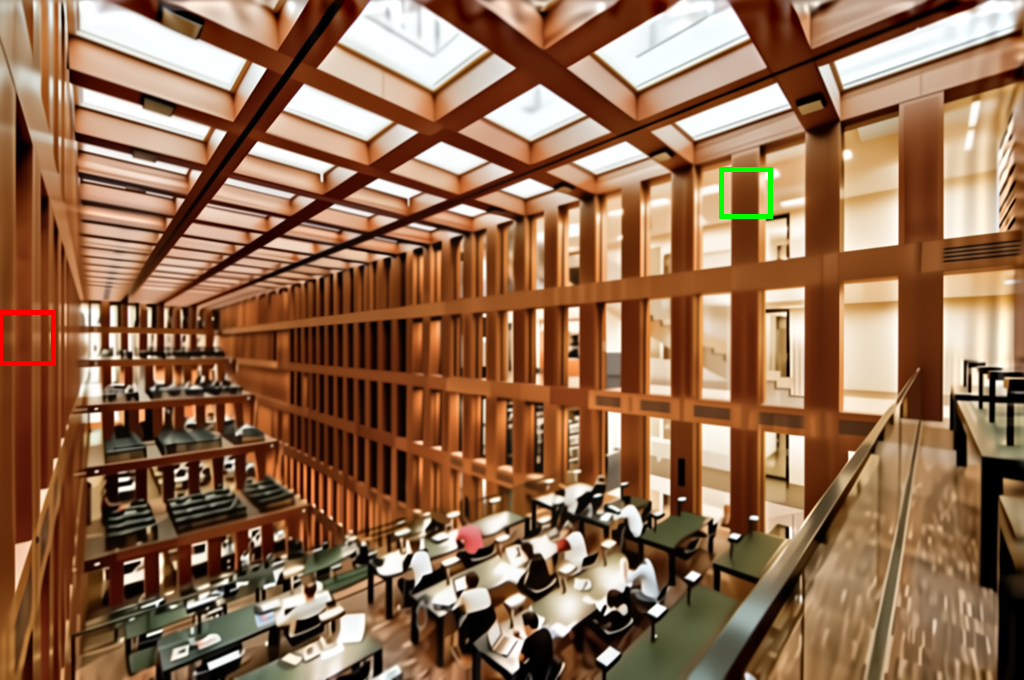}
      \rb{\includegraphics[width=0.499\linewidth, height=0.499\linewidth]{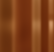}}\hfill
  	  \gb{\includegraphics[width=0.499\linewidth, height=0.499\linewidth]{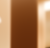}}\hfill
    \end{minipage}
	}
  \subfloat[DSDNet (ours)]{
    \begin{minipage}{0.195\linewidth} 
      \centering
      \includegraphics[width=\linewidth]{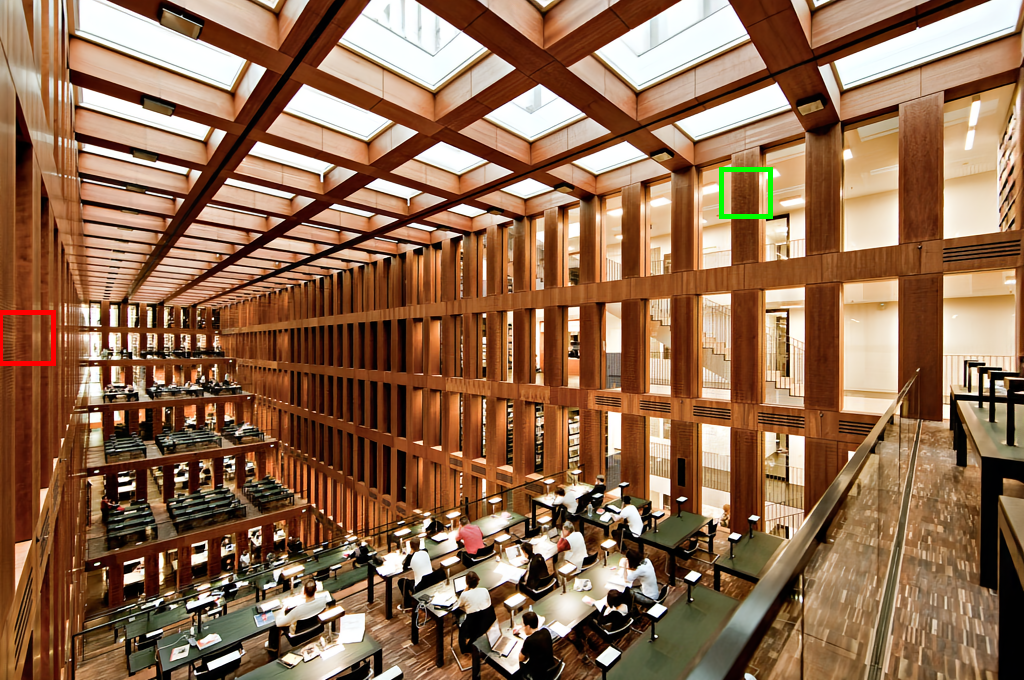}
      \rb{\includegraphics[width=0.499\linewidth, height=0.499\linewidth]{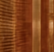}}\hfill
  	  \gb{\includegraphics[width=0.499\linewidth, height=0.499\linewidth]{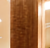}}\hfill
    \end{minipage}
	}
  \subfloat[Ground truth]{
    \begin{minipage}{0.195\linewidth} 
      \centering
      \includegraphics[width=\linewidth]{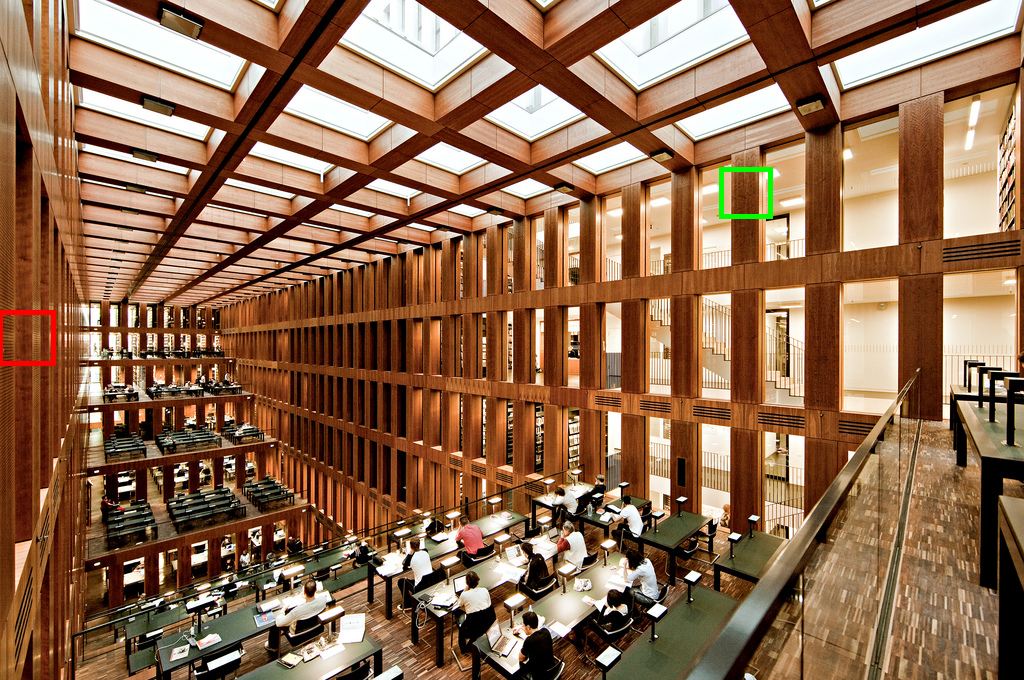}
      \rb{\includegraphics[width=0.499\linewidth, height=0.499\linewidth]{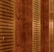}}\hfill
  	  \gb{\includegraphics[width=0.499\linewidth, height=0.499\linewidth]{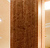}}\hfill
    \end{minipage}
	}\hfill				
  \end{minipage}
	\vfill
	
      \caption{A synthetic case comes from \textsc{Lai} \cite{lai2016comparative}. Our method restores clearer images with finer details (e.g., the wood texture).}
\label{fig:ev_2}
\end{figure*}

\begin{figure*}[!h]
\raggedright
  \begin{minipage}[b]{.95\linewidth}
  \subfloat[Blurry input]{
    \begin{minipage}{0.243\linewidth} 
      \centering
      \includegraphics[width=\linewidth]{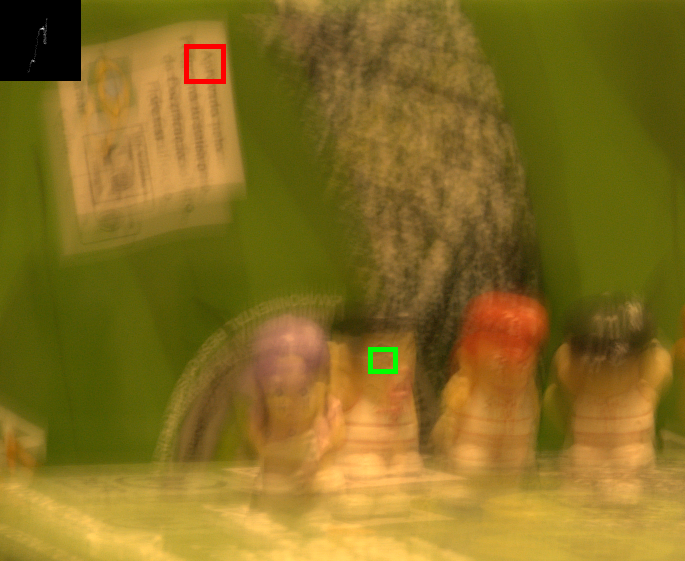}
      \rb{\includegraphics[width=0.499\linewidth, height=0.499\linewidth]{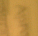}}\hfill
  	  \gb{\includegraphics[width=0.499\linewidth, height=0.499\linewidth]{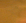}}\hfill
    \end{minipage}
	}
  \subfloat[IRCNN \cite{zhang2017learning}]{
    \begin{minipage}{0.243\linewidth} 
      \centering
      \includegraphics[width=\linewidth]{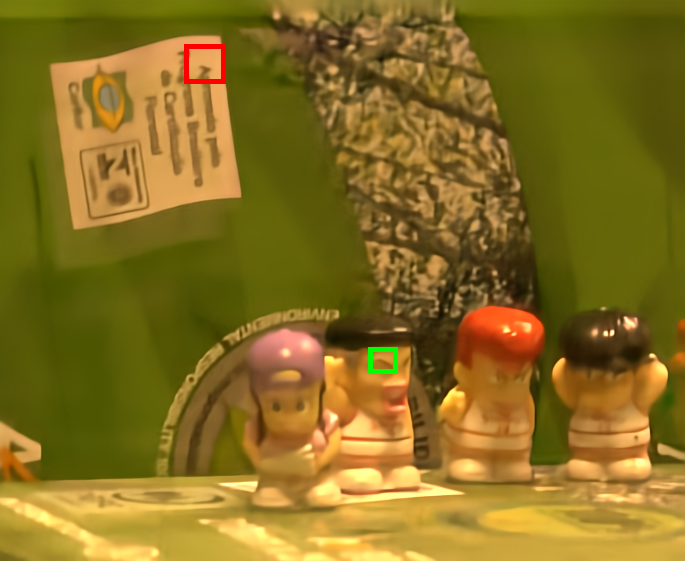}
      \rb{\includegraphics[width=0.499\linewidth, height=0.499\linewidth]{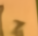}}\hfill
  	  \gb{\includegraphics[width=0.499\linewidth, height=0.499\linewidth]{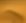}}\hfill
    \end{minipage}
	}
  \subfloat[ADM\_UDM \cite{ko2020deep}]{
    \begin{minipage}{0.243\linewidth} 
      \centering
      \includegraphics[width=\linewidth]{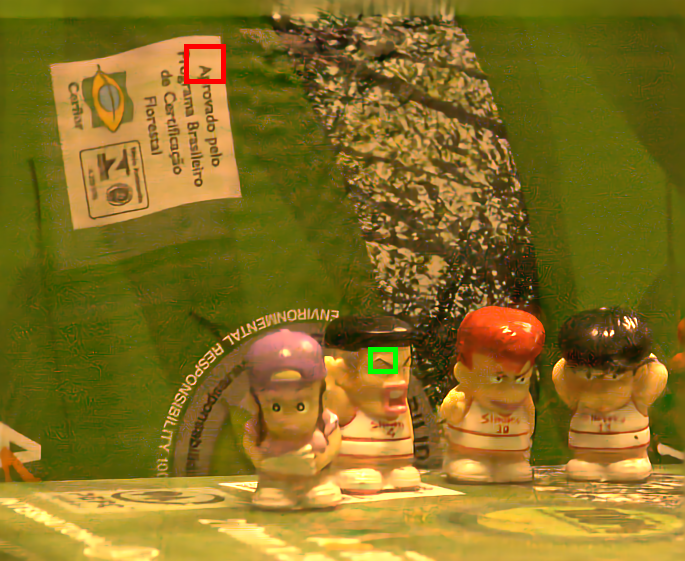}
      \rb{\includegraphics[width=0.499\linewidth, height=0.499\linewidth]{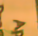}}\hfill
  	  \gb{\includegraphics[width=0.499\linewidth, height=0.499\linewidth]{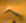}}\hfill
    \end{minipage}
	}
  \subfloat[KerUNC \cite{nan2020deep}]{
    \begin{minipage}{0.243\linewidth} 
      \centering
      \includegraphics[width=\linewidth]{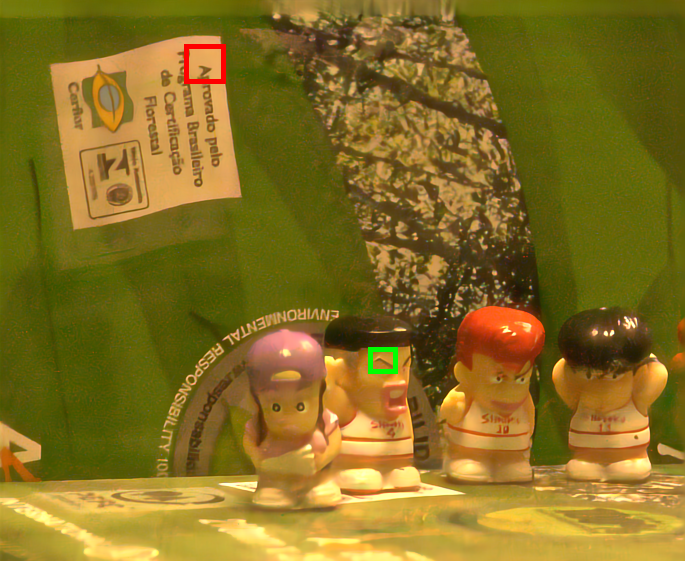}
      \rb{\includegraphics[width=0.499\linewidth, height=0.499\linewidth]{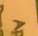}}\hfill
  	  \gb{\includegraphics[width=0.499\linewidth, height=0.499\linewidth]{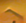}}\hfill
    \end{minipage}
	} \\
  \subfloat[DWDN \cite{dong2020deep}]{
    \begin{minipage}{0.243\linewidth} 
      \centering
      \includegraphics[width=\linewidth]{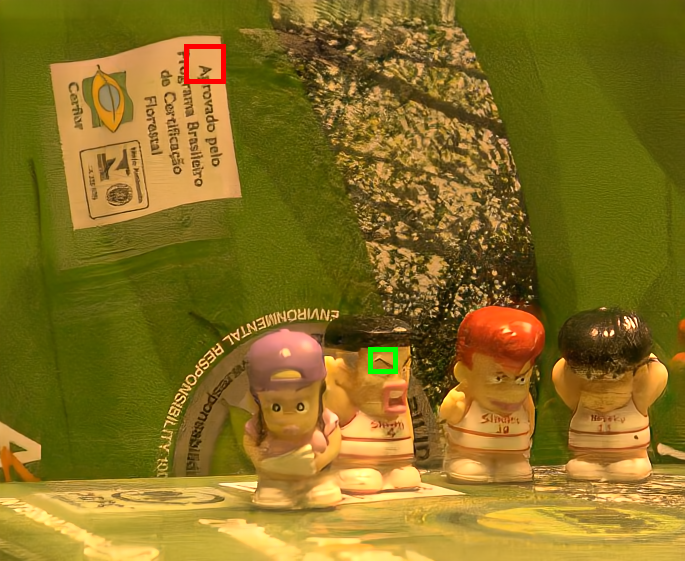}
      \rb{\includegraphics[width=0.499\linewidth, height=0.499\linewidth]{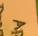}}\hfill
  	  \gb{\includegraphics[width=0.499\linewidth, height=0.499\linewidth]{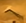}}\hfill
    \end{minipage}
	}
  \subfloat[SVMAP \cite{dong2021learning}]{
    \begin{minipage}{0.243\linewidth} 
      \centering
      \includegraphics[width=\linewidth]{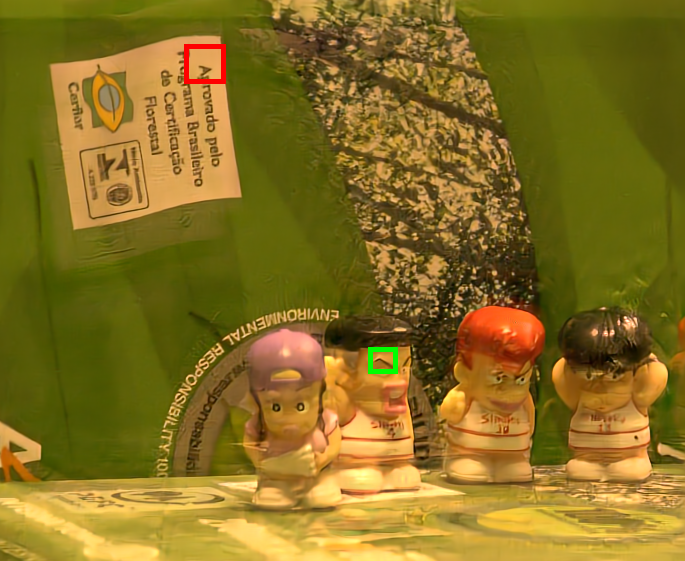}
      \rb{\includegraphics[width=0.499\linewidth, height=0.499\linewidth]{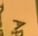}}\hfill
  	  \gb{\includegraphics[width=0.499\linewidth, height=0.499\linewidth]{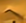}}\hfill
    \end{minipage}
	}
  \subfloat[DRUNet \cite{zhang2021plug}]{
    \begin{minipage}{0.243\linewidth} 
      \centering
      \includegraphics[width=\linewidth]{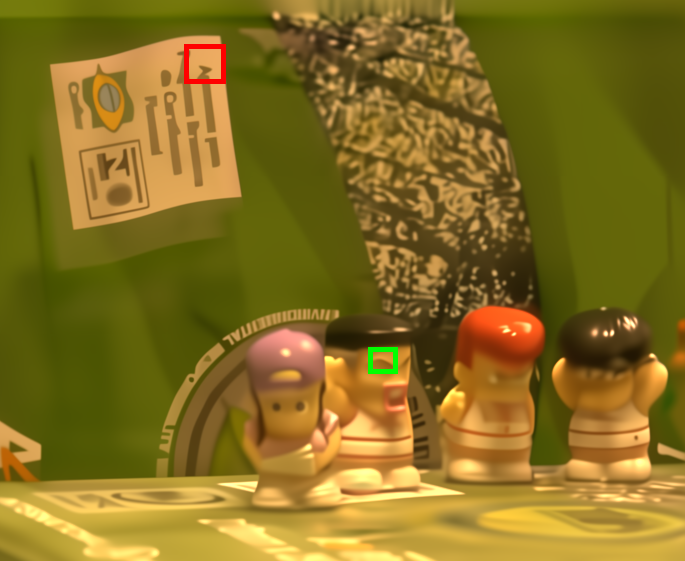}
      \rb{\includegraphics[width=0.499\linewidth, height=0.499\linewidth]{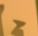}}\hfill
  	  \gb{\includegraphics[width=0.499\linewidth, height=0.499\linewidth]{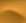}}\hfill
    \end{minipage}
	}
  \subfloat[DSDNet (ours)]{
    \begin{minipage}{0.243\linewidth} 
      \centering
      \includegraphics[width=\linewidth]{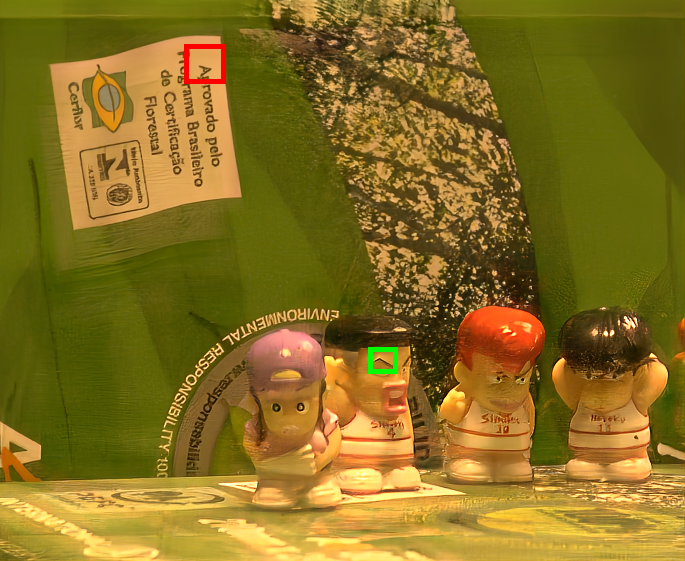}
      \rb{\includegraphics[width=0.499\linewidth, height=0.499\linewidth]{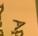}}\hfill
  	  \gb{\includegraphics[width=0.499\linewidth, height=0.499\linewidth]{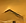}}\hfill
    \end{minipage}
	}\hfill	
  \end{minipage}
	\vfill
      \caption{A real-world case whose kernel is estimated by \cite{pan2016blind}. Our method generates a better image with clearer text and a sharper eyebrow. In the olive background, our method produces the sharpest result without noise.}
      \label{fig:ev_3}
\end{figure*}

We show visual comparisons of a synthesized and a real-world case in Fig.~\ref{fig:ev_2} Fig.~\ref{fig:ev_3}, respectively. 

Fig.~\ref{fig:ev_2} shows the results of \textsc{Manmade} from the dataset \textsc{Lai} \cite{lai2016comparative}. The evaluated methods generate blur results. In contrast, our method reconstructs better images (e.g., the wood texture is better restored, as shown in both red and green boxes).
%
We show some visual comparisons on BSD100, \textsc{Lai} and a real-world case in Fig.~\ref{fig:ev1}, Fig.~\ref{fig:ev_2} and Fig.~\ref{fig:ev_3}, respectively.

Fig.~\ref{fig:ev1} shows the deblurred results of one synthetic blurry image with $1\%$ Gaussian noise. Our method generates the clearer image with finer details as shown in the parts enclosed in the red and green boxes. 

Fig.~\ref{fig:ev_2} shows the results of \textsc{Manmade} from the dataset \textsc{Lai} \cite{lai2016comparative}. The evaluated methods generate the results with blur effect. In contrast, our method reconstructs better images (e.g., the wood texture is better restored as shown in both red and green boxes).

Fig.~\ref{fig:ev_3} shows the deblurred results of a pair of real-world blurry images and estimated kernel \cite{pan2016blind}.
Our method restores the text to sharpness in the red boxes and the blackest and sharpest eyebrow in the green boxes.
%
%
In contrast, other methods cannot restore the text well and mix the eyebrow with the skin color. 
Furthermore, they also generate artifacts and noise in the olive background.

\subsection{Ablation Study}

\label{subsec:ablation}
\begin{table}[ht]
\caption{Ablation study on \textsc{Set5}. ``w/o \textbf{F}, \textbf{G}" is out of the convolutional layers before the Maxout layers; ``ReLU" and ``RBF" replace the Maxout layers with ReLU and RBF layers, respectively; ``CG" performs conventional Conjugate Gradient method for deconvolution rather than CGNet, ``FFT" performs FFT deconvolution with edge taper \cite{kruse2017learning}, and the superscript $^\dagger$ means the input is denoised by DRUNet \cite{zhang2021plug} first.}
\label{tb:ablation}
\centering
\scriptsize
\tabcolsep=4pt
\begin{tabular}{l c c c c c c c c}
\toprule
~ &w/o $\mathbf{F}$, $\mathbf{G}$ &ReLU &RBF &CG  &CG$^\dagger$ &FFT &FFT$^\dagger$  &DSDNet \\ \midrule
PSNR(dB)         &26.66  &32.78  &32.98  &29.07  &32.39  &31.30  &32.03   &33.11\\
FLOPs(G)         &136.26 &464.94 &468.03 &470.73 &559.97 &288.89 &391.57  &466.32\\
Parameters(M)    &1.04   &237.85 &237.85 &0.31   &32.95  &0.27   &32.09   &237.87\\ \midrule
Gain(dB)         &-6.45  &-0.33  &-0.13  &-4.04  &-0.72  &-1.81  &-1.08   &-/-\\
\bottomrule
\end{tabular}

\end{table}

In this section, we design experiments to show the efficiency of the proposed discriminative shrinkage functions and the differentiable CGNet.
Table~\ref{tb:ablation} shows the ablation results w.r.t. different baselines.
In this study, we train 7 models based on the architecture of the Heavy DSDNet.
We compare the number of floating point operations (FLOPs) and the parameters in this study.

To validate the effects of the $\mathbf{F}_i$ and $\mathbf{G}_j$, we train a model without estimating these two filters, denoted by ``w/o $\mathbf{F}$,$\mathbf{G}$".
Without the feature maps coming from them, we can only learn the shrinkage functions from RGB inputs.
Table~\ref{tb:ablation} shows that the PSNR value of the results by the baseline method without $\mathbf{F}_i$ and $\mathbf{G}_j$ is at least 6.45dB lower than that of our approach.

We also evaluate the effect of the Maxout layers by replacing them with ReLU. 
Table~\ref{tb:ablation} shows that the method using ReLU does not generate better results than the proposed method, suggesting the effectiveness of the Maxout layers.

As RBFs are usually used to approximate shrinkage functions, one may wonder whether using them generates better results or not. To answer this question, we replace the Maxout layers with the commonly-used Gaussian RBFs. Table~\ref{tb:ablation} shows that the PSNR value of the method using RBFs is at least 0.13dB lower than that of our method, indicating the effectiveness of the proposed method.  

%
%
%
Finally, to demonstrate the efficiency of the proposed CGNet, we train 2 models with the conventional CG method and 2 models with FFT deconvolution.
%
%
%
The CG method is unstable with respect to even small perturbations \cite{marin2001conjugate,hadamard2003lectures}. Each optimization step in training and the existence of noise may cause the divergence of the CG method.
Hence we have to reduce the learning rate and apply gradient clipping to avoid the gradient exploding during the training.
However, it still performs poorly even if the training can be finished.
To make the training more feasible, we denoise the inputs first by DRUNet \cite{zhang2021plug}, and this model is denoted as ``CG$^\dagger$".
The training of ``CG$^\dagger$" is smoother, the learning rate can be set as that of the DSDNet, and the performance is much better than generic ``CG".
With another model to denoise, the computational cost is about 26\% more FLOPs, and it takes more than twice the time for training compared to the DSDNet.

Similar to CG ones, we also provide results of deconvolution via FFT with the artifact processing operation by \cite{kruse2017learning}, i.e., ``FFT" and ``FFT$^\dagger$". Although FFT ones are much faster than CG ones, the gap between ``CG$^\dagger$" and ``FFT$^\dagger$" is considerable, as mentioned in Section~\ref{sec: revisting}.
As for the test time on \textsc{Set5}, ``CG" is 5.6001 seconds, ``CG$^\dagger$" is 5.8252 seconds, ``FFT" is 3.6941 seconds, ``FFT$^\dagger$" is 4.0330 seconds, and DSDNet is 2.6717 seconds.
These result show the efficiency of the proposed CGNet. 
We include the ablation study on HypNet and NLNet in the supplementary material.
\begin{figure}[t]
    \centering
    \includegraphics[width=.5\linewidth, keepaspectratio]{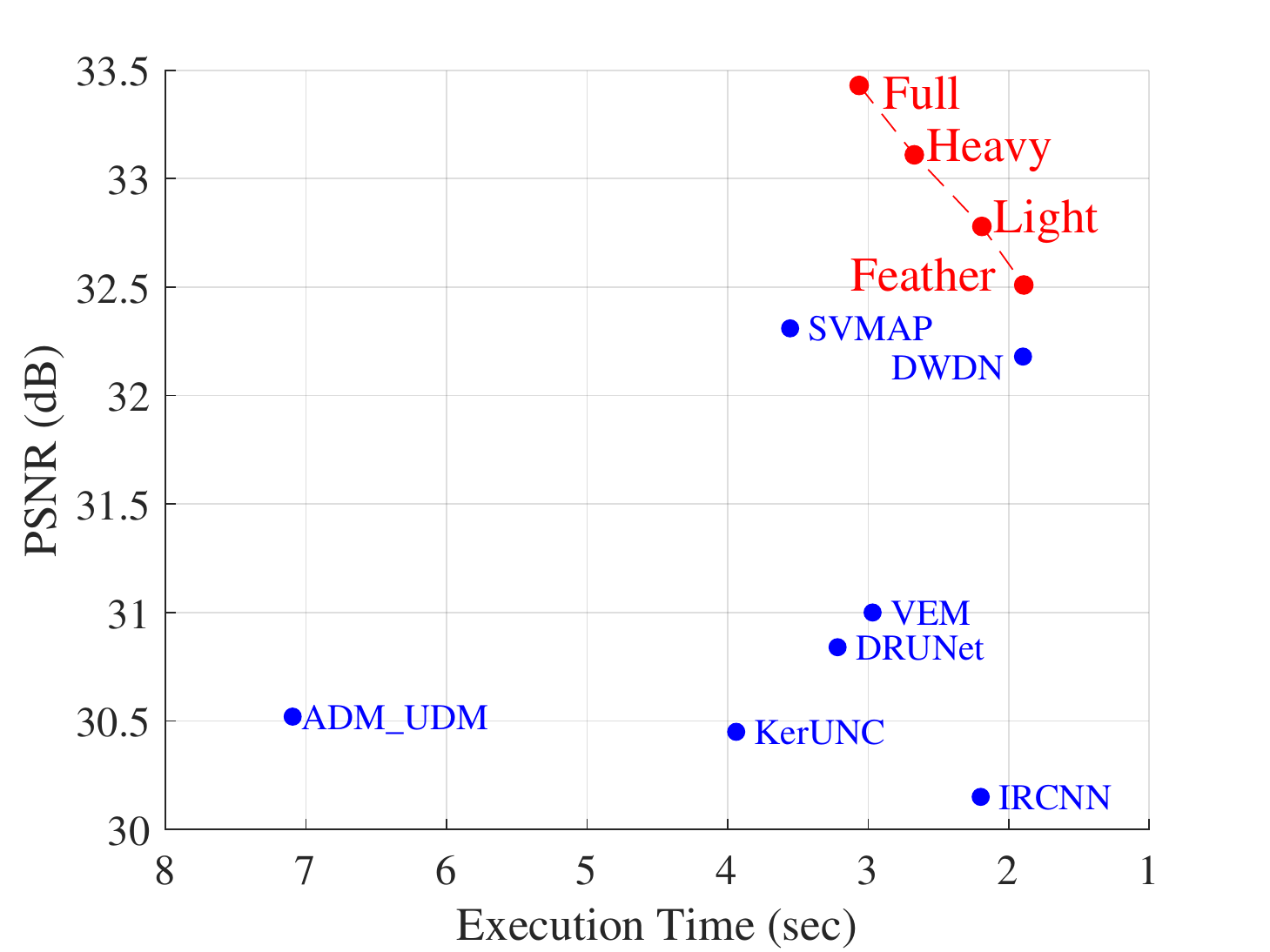}
    \caption{Speed and accuracy trade-off. The results are evaluated on the \textsc{Set5} dataset with $1\%$ Gaussian noise. Results of this work are shown in red points; the fastest Feather DSDNet still performs favorably against the state-of-the-art methods (blue points) in PSNR.} 
    \label{fig:time}
\end{figure}
\subsection{Execution Time Analysis}

\label{subsec:time}
We analyze the execution time of the proposed methods and state-of-the-art ones. 
All the execution times are evaluated on one nVidia RTX 2080Ti GPU.
Fig.~\ref{fig:time} shows that our models are faster and more accurate than the state-of-the-art methods.
Among these methods, our Feather model is a little bit faster than DWDN by 0.0052 seconds yet outperforms all other methods in PSNR.

\subsection{Limitations}

\label{subsec:limit}
Although better performance on various datasets has been achieved, our method has some limitations. Our model cannot deal with blurry images containing significant saturation regions, which may lead to overflow. More analysis can be found in the supplementary material. 
%
%
%

%
%

\section{Conclusion}

\label{sec:conclusion}
In this paper, we present a fully learnable MAP model for non-blind deconvolution. We formulate the data and regularization terms as the learnable ones and split the deconvolution model into data-related and regularization-related sub-problems in the ADMM framework. Maxout layers are used to learn the discriminative shrinkage functions, which directly approximate the solutions of these two sub-problems. We have further developed a CGNet to restore the images effectively and efficiently. With a reasonable design, the size of our model is flexibly adjustable while keeping competitive in performance. Extensive evaluations on the benchmark datasets demonstrate that the proposed model performs favorably against the state-of-the-art non-blind deconvolution methods in terms of quantitative metrics, visual quality, and computational efficiency.


\clearpage
%
%
\bibliographystyle{splncs04}
\bibliography{egbib}
\end{document}